\documentclass[pdflatex,sn-mathphys-ay]{sn-jnl}

\usepackage{rotating} 
\usepackage[T1]{fontenc}
\usepackage{makecell}
\usepackage{quoting}
\quotingsetup{leftmargin=0pt, rightmargin=0pt}
\usepackage{framed}
\usepackage{soul}
\usepackage{array}
\usepackage{multirow}
\usepackage{balance}

\usepackage{graphicx}%
\usepackage{amsmath,amssymb,amsfonts}%
\usepackage{amsthm}%
\usepackage{mathrsfs}%
\usepackage[title]{appendix}%
\usepackage{xcolor}%
\usepackage{textcomp}%
\usepackage{manyfoot}%
\usepackage{booktabs}%
\usepackage{algorithm}%
\usepackage{algorithmicx}%
\usepackage{algpseudocode}%
\usepackage{listings}%

\theoremstyle{thmstyleone}%
\theoremstyle{thmstyletwo}%

\theoremstyle{thmstylethree}%

\begin{document}

\title{Assessing and Explaining the Persuadability of Large Language Models as Legal Decision Tools}


\author[1]{\fnm{Oisin} \sur{Suttle}} 
    \email{oisin.suttle@mu.ie}

\author[2]{\fnm{David} \sur{Lillis}} 
    \email{david.lillis@ucd.ie}

\affil[1]{
\orgdiv{School of Law and Criminology}, \orgname{Maynooth University}, \orgaddress{Maynooth, Ireland}}

\affil[2]{
\orgdiv{School of Computer Science}, \orgname{University College Dublin},  \orgaddress{Dublin, Ireland}}

\abstract{As Large Language Models (LLMs) are proposed as legal decision assistants, and even first-instance decision-makers, across a range of judicial and administrative contexts, it becomes essential to explore how they answer legal questions, and in particular the factors that lead them to decide difficult questions. A specific feature of legal decisions is the need to respond to arguments advanced by contending parties. A legal decision-maker must be able to engage with, and respond to, including through being potentially persuaded by, these arguments. Conversely, they should not be unduly persuadable, deciding cases based on the skills of the advocates rather than the merits of the case. In this paper we explore how frontier open- and closed-weights LLMs respond to legal arguments. We propose a metric to measure persuadability in the trilateral setting in which competing advocates seek to persuade a judge of opposite conclusions. We report original experimental results measuring how far the quality of the advocate making arguments affects the likelihood that a given model will agree with a particular legal point of view. We further examine how far models are capable of distinguishing between stronger and weaker arguments and how far model judgments in this domain are affected by positional bias. Through parallel bilateral trials we show how the trilateral setting changes the demands on judge models, and in turn their apparent persuadability. Finally we examine the specific features of arguments that affect persuasion, including the relative contribution of legal content and rhetorical form, the extent to which model persuasion tracks human expert judgments of argument quality, and the extent to which argument quantity, diversity and type affect persuasive outcomes. Our results have implications for the feasibility of adopting LLMs across legal and administrative settings.}


\maketitle

\section{Introduction}

As Large Language Models (LLMs) are proposed as legal decision assistants, and even decision-makers, across a range of judicial and administrative contexts~\citep{Volokh2019,Gutierrez2024a,Lai2024,Pesch2025,Mamalis2024,He2024}, it becomes essential to explore how these models answer legal questions, and in particular the factors that lead them to decide difficult questions in one way or another\footnote{An earlier version of this work was accepted for the Proceedings of the 21st International Conference on Artificial Intelligence and Law (ICAIL 2026), DOI: \href{https://doi.org/10.1145/3836937.3837003}{10.1145/3836937.3837003}; arXiv preprint: \href{https://arxiv.org/abs/2604.26233v2}{arXiv:2604.26233v2}. The present manuscript substantially extends that version.}. One specific feature of legal decisions is the need to respond to arguments advanced by contending parties. A legal decision-maker must be able to engage with, and respond to, including through being potentially persuaded by, arguments advanced by the parties. Conversely, they should not be unduly persuadable, influenced by a particularly compelling advocate to decide cases based on the skills of the advocates, rather than the merits of the case. This paper explores how frontier open- and closed-weights LLMs respond to legal arguments, including examining how far the quality of the advocate making those arguments affects the likelihood that a model will agree with a particular legal point of view and examining the factors explaining model persuadability, including examining the relative role of legal content and rhetorical form, the significant differences between bilateral and trilateral persuasion settings, stability given variation in argument order, conformity of model persuasion with human judgment, and the significance of argument type, diversity and quantity in explaining persuasive outcomes.

It is a fundamental principle of natural justice that the subject of a judicial or administrative decision should have the opportunity to be heard in relation to that decision, and to put forward arguments as to how that decision should be made~\citep{Waldron2023}. This is expressed in the Latin maxim \emph{audi alteram partem}, and is a fundamental principle of administrative law and due process. \citet[p.~73]{Cane2011} characterises the administrative right to a fair hearing as comprising \emph{inter alia} ``the right to present one's case orally or in writing or both, the right to examine and cross-examine witnesses (including one's opponent), the right to be represented (perhaps by a qualified lawyer), the right to have a decision based solely on material which has been available to (and so answerable by) the parties, the right to a reasoned decision which takes proper account of the evidence and addresses the parties' arguments''. In modern practice, this is typically regarded as a requirement of fairness or justice: it is a right of each party to have their case heard. By contrast, in its classical roots the principle equally expressed the concern that a decision-maker was more likely to get the answer right having heard from all parties~\citep{Kelly1964}. Regardless of how it is understood, the principle requires not only that a party be heard, but also that the decision-maker be open to being persuaded by what they hear: a decision-maker who has prejudged the question, reaching a conclusion before hearing from one of the parties, cannot provide a fair hearing. The requirement to hear from all parties requires at least some possibility of being persuaded by all parties.

However, the requirement that a decision-maker be persuadable does not obviate the need for the decision-maker to themselves make the decision, and for that decision to be an exercise of the decision-maker's own judgment. The decision-maker is not simply a transmission mechanism for the arguments of the parties. Rather, we appoint as responsible decision-makers persons who we think, for whatever reason, are likely to make good decisions, and we expect those decisions to be theirs~\citep{Solum2003}. Judges thus require ``intellectual autonomy, to form independent views about the case, rather than being unduly subjected to the influence of lawyers or other judges''~\citep{Amaya2025}. In many legal contexts, there are no clear objective standards that fully determine the right decision: either legal answer may be logically defensible, so we entrust the decision-maker to decide the question on its own merits~\citep{Gardner2001}. We expect them to treat the parties fairly, by hearing their side of the story, and to do all they reasonably can to find the true, right or best answer, including by considering the arguments of the parties; but the final decision will be theirs, responsive to but not determined by those arguments.

There is thus a tension in the judicial role: the judge must be persuadable, but not unduly so. A judge who never changes their view in response to a compelling argument is a bad judge; but so is a judge who always agrees with the most recent argument they have heard, or the argument put forward by the most convincing advocate. A good judge must exhibit ``the willingness to modify one's position in light of other people's arguments, in a way that avoids both floppiness and fleetingness, on the one hand, and rigidity and stubbornness, on the other''~\citep{Amaya2025}. If LLMs are to serve in legal decision-support roles, it is therefore important to understand how they respond to this tension.

In this paper we investigate the persuadability of LLMs in response to a set of `hard' legal questions, in respect of which legal experts are likely to disagree, across three jurisdictions (the United States, England and Wales, and Ireland). We focus on hard questions because it is in respect of these that persuasion is most significant. On straightforward questions, where there is one right answer, we expect competent decision-makers to consistently identify that answer, independent of any arguments advanced by the parties. Further, in relation to easy questions, we can evaluate persuasion by whether it leads decision-makers to more frequently find that right answer; but in respect of hard questions, there is no ground truth against which we can test the answers reached by decision-makers, requiring alternative methods for identifying persuasion and its causes. In our experiments, we focus on the ways decisions vary depending on the quality of the advocate on either side. By identifying a number of advocates who are typically more or less convincing, and asking our decision-makers to consider arguments authored by each, we can identify how far decisions are a function of the underlying laws and facts, and how far they reflect the arguments advanced by those advocates, and hence the persuadability of the relevant decision-maker. This allows us, first, to measure the persuadability of models in the face of competing arguments, and second, to explore how various features of both arguments and set-up affect persuasion outcomes.

Section~\ref{sec:related} outlines how our work relates to existing scholarship in four principal areas: on LLMs and legal reasoning; on LLMs as agents of persuasion; on LLMs as targets of persuasion; and on legal argument mining. Section~\ref{sec:approach} explains our experimental approach. Section~\ref{sec:metrics} explains the metrics that we adopt to measure and compare persuadability across models.

Our results are presented in Section~\ref{sec:results}. Initially in Section~\ref{sec:results:overall}, we show the overall results of our primary experiment, which investigates the degree to which LLMs are persuadable when used as legal decision-makers in a trilateral setting where two competing Advocate entities vie to persuade a Judge entity of the merits of their respective arguments. The findings in this section prompt further investigation into the factors that may influence this persuadability. Section~\ref{sec:results:scenario} examines the results at the more granular level of individual scenarios (i.e. cases), to ascertain whether an LLM judge has significant systematic preferences for particular parties. Following this, we investigate the effect of positional bias by varying the order in which arguments are presented (Section~\ref{sec:results:order}). Section~\ref{sec:results:bilateral} then contrasts our trilateral setup with the more traditional bilateral persuasion configuration (where a single party tries to persuade another), which also provides further insights into the behaviour observed in the earlier results.

Section~\ref{sec:results:features} outlines a number of additional experiments to investigate the features of persuasive arguments that potentially have an effect. Section~\ref{sec:results:content-vs-form} presents approaches to investigating whether it is the legal content of arguments or the rhetorical form in which they are presented that has a greater effect on persuasiveness; Section~\ref{sec:results:human} compares the results of human expert assessments of argument strength with those of the LLM judges; and Section~\ref{sec:results:quantity-type-diversity} investigates whether the quantity and type of the arguments presented has an effect on the models used to decide between them, using an argument typology approach.

Section~\ref{sec:future} identifies further implications and directions for future research.

We make the following key contributions:

\begin{itemize}
    \item We identify and explore a key tension in the use of LLMs as administrative or judicial decision supports, namely that a decision-maker should be open to persuasion by compelling legal arguments, but not to an excessive degree that renders them overly impressionable.
    \item We propose a novel method of measuring the persuadability of LLMs in trilateral settings and demonstrate the distinct features of trilateral as against bilateral persuadability.
    \item We provide the first systematic exploration of the persuadability of LLMs in legal settings, including measuring the persuadability of a range of open- and closed-weights models and exploring the relevance of legal content and rhetorical form, the role of argument order, and the impact of argument quantity, type and diversity in shaping model persuadability.
\end{itemize}

All of the code and data used in this study is publicly available, including the templates used for generating prompts and the prompts themselves\footnote{\url{https://anonymous.4open.science/r/AIL-2026/}}. 


Part of the work presented in this paper has previously been published as a short conference paper in~\citep{Suttle2026}, specifically the initial overall results from Section~\ref{sec:results:overall} and some early work on distinguishing between the effects of legal substance versus argumentative form (contained in greatly expanded form in Section~\ref{sec:results:content-vs-form}). The other results and analyses are new to this paper, including analysis relating to the ordering of parties in the prompted scenarios, contrasting between trilateral and bilateral argumentation scenarios, assessment of argument strength with human assessors and further analysis of the persuasive power of arguments according to their quantity, type and diversity. 

\section{Existing Research} \label{sec:related}

\subsection{LLMs and Legal and Normative Reasoning}
Various studies have tested the capacity of LLMs for legal and moral reasoning. These have included studies examining the psychological structure~\citep{Almeida2024} and values~\citep{Jiao2025} implicit in LLMs' moral and political judgments, and how these can be aligned~\citep{Tennant2025}. Other work has specifically addressed LLMs' capacities to analyse and resolve legal questions~\citep{Guha2023}, examining both their accuracy~\citep{Posner2025} and stability~\citep{Blair-Stanek2026}. Studies have also examined LLMs' competence in specific forms of legal reasoning, including statutory identification~\citep{Surani2025} and interpretation~\citep{Blair-Stanek2023}, constitutional interpretation~\citep{Coan2025}, syllogistic reasoning~\citep{Zhang2025} and purposive reasoning~\citep{Nunes2025}. Work on LLMs as legal judgment predictors is also relevant here~\citep{Wei2025}. More generally, legal-theoretical work on AIs as legal decision-makers has clarified the normative expectations of judges, and helped clarify whether and under what circumstances we should consider LLMs as permissible tools for legal decision-making~\citep{Volokh2019,Tasioulas2023}.

\subsection{LLMs as agents of persuasion}
There is now an extensive literature on LLMs as persuaders, predominantly focussed on the effectiveness of LLMs in persuading human counterparties on a range of factual, moral and political issues~\citep{Durmus2024,Carrasco-Farre2024,Breum2024,OpenAI2024,Rogiers2024,Jones2024,Schoenegger2025}. Much of this literature has been motivated by concerns about the use of LLMs for misinformation and has in consequence focussed on persuasion in informal settings and on questions of politics and public policy (e.g. public health), where persuasion may be most salient, and factual questions, in order to model misinformation~\citep{Bozdag2025}. Other work explores the capacity and willingness of LLMs to adopt various (including unethical) persuasion strategies or goals, and to exploit addressee characteristics~\citep{Liu2025,Hackenburg2024,Ju2025}. Others have examined how knowledge that an argument was AI generated affected its persuasiveness~\citep{Teigen2024}. These studies have identified model size as an important predictor of effective persuasion~\citep{Durmus2024,Idziejczak2025}, although this may be subject to a ceiling effect~\citep{Hackenburg2024}. Model architecture, specifically reasoning architecture, is also identified as predicting persuasion effectiveness~\citep{Zhao2025}.

\subsection{LLMs as objects of persuasion}
While most work on LLMs and persuasion has focussed on the LLM as persuader, research has also examined the complementary question of the persuadability of LLMs~\citep{Ju2025,Zeng2024}. This has included efforts to identify strategies to enable LLMs to distinguish good arguments, which tend towards truth, from misleading arguments, in the context of factual issues where there is a relevant ground truth~\citep{Tan2025}. Prior work has also identified variation in persuadability depending on the subject matter/domain~\citep{Tan2025}, and explored whether specific types/styles of argument are more or less successful~\citep{Tan2025}. Model size and reasoning architecture have been identified as relevant to persuadability, with larger and reasoning models less readily persuaded~\citep{Zhao2025}. As with work on LLMs as persuaders, with which it overlaps, this work has typically focussed on either open moral or policy questions, or factual questions, and bilateral rather than trilateral scenarios~\citep{Tan2025}.

\subsection{Argumentation Theory and Argument Mining}
There is an extensive literature on argument theory and argument mining. (For a review in the legal context, see~\citet{Zhang2022a}) This has included work developing both general purpose and domain specific argument typologies and schemes. In the legal domain, this work has explored the automated detection of argument components and relations (e.g. \citet{Stab2014, Stab2017}). While much of this literature has relied on generic argument schemas (e.g. \citet{Poudyal2020}), more recent work has sought to distinguish among specific types of legal argument across settings including the European Court of Human Rights \citep{Habernal2024}, German Constitutional Court \citep{Luders2025}, Italian Court of Cassation \citep{Giacchetta2025}, Court of Justice of the European Union \citep{Grundler2022a}, and Czech Supreme Court \citep{Dugac2025}. In the legal context, this work has typically focussed on mapping arguments in judicial opinions, but has not generally addressed causal questions about how argument type affects outcomes.

These literatures inform our study, both contributing hypotheses and identifying important gaps that we seek to fill. First, while there is evidence that persuasion varies across domains, existing work on LLM persuasion has not addressed legal questions. The legal domain combines factuality (there sometimes are objectively right answers) with subjective judgment (there often are not). In many cases, there is no ground truth, but there are explicit norms and criteria that must be accommodated in any satisfactory argument, making the task of persuasion potentially more complicated than other domains. Second, existing work has focussed predominantly on persuasive dialogues between persuader and persuadee. By contrast, the legal domain is most often trilateral, with two or more persuaders seeking to influence a third-party decision-maker in opposite directions. However, the task of the decision-maker is to adjudicate the substantive question, rather than to judge the competing arguments, distinguishing their role from that of a debate judge. \citet{He2024} propose a judicial support model that includes party debate, but do not evaluate the impact of argument quality. Existing results identifying the importance of model size and reasoning architecture, and variation in persuadability across contexts, inform specific features of our experimental design below. We draw on argument mining approaches in our supplemental analysis to identify the content of arguments and how this affects model persuasion.

\section{Experimental Approach} \label{sec:approach}

Our experimental approach can be summarised briefly. We first identify a suitable set of hard legal questions. We next use a set of Advocate models to generate a range of arguments, of varying qualities, for each side of these hard questions. Given this bank of scenarios and arguments, we prompt a range of Judge models, giving them the relevant facts and a selection of arguments. By measuring how Judge models' responses vary given different arguments, we can identify and compare these models' persuadability, and in turn examine the features of arguments and argumentative setting that affect model persuasion.

\subsection{Dataset: Identifying Hard Legal Questions}

Our first task is to identify a suitable set of such `hard' questions. A hard question, for our purposes, is one about which competent legal experts are likely to disagree, with the consequence that there is no uncontroversial ground truth against which model answers can be judged. We follow~\citet{Blair-Stanek2026} in using appellate court split decisions, where there is at least one dissenting opinion, as a proxy for identifying these hard questions. We assume that appellate judges are competent legal experts motivated to provide the best answer to questions presented to them, so the existence of a dissenting opinion in an appellate case means that competent experts reached different conclusions on the question posed.

For our US scenarios, we draw a random selection from Blair-Stanek and Van Durme's dataset of summaries of US Court of Appeals split decisions\footnote{\url{https://github.com/BlairStanek/legal_instability}}. For our samples from England and Wales (E\&W) and from Ireland, we follow their procedure for generating summaries, presenting the full text of all judicial opinions in a case (majority and dissenting) to the OpenAI o1 model, and prompting it to generate a summary of both the facts and the legal arguments\footnote{Our prompt is identical to Blair-Stanek and Van Durme, with minor amendments to reflect specific national features. E.g. different examples of state agencies, different court names etc. While o1 is no longer a state-of-the-art model, we chose to prioritise consistency across our dataset over using a more recent model.}. For England and Wales, we take the five most recent (prior to October 2025) split decisions of the Court of Appeal for England and Wales. For our Irish sample, we used the five most recent split decisions of the Irish Supreme Court. This results in a total of 15 scenarios: 5 per jurisdiction. Our case summaries (again, following Blair-Stanek and Van Durme) comprise three paragraphs setting out the relevant facts, and two describing the principal legal arguments advanced by the two sides in the case.

Our choice to draw scenarios from different anglophone jurisdictions reflects expected differences in the capacities of both our Judge and Advocate models. Existing studies have identified variance in LLMs' embedded knowledge by jurisdiction~\citep{Dahl2024}. In particular, LLMs appear to have greater knowledge of US law than of other anglophone legal systems, presumably reflecting the larger corpus of US legal materials in their training data~\citep{Curran2025}. By drawing scenarios from the US, but also from one mid-sized (E\&W) and one smaller (Ireland) jurisdiction, we can examine whether and how far embedded legal knowledge affects persuadability. We might expect that a model would be less persuadable in relation to a question where its embedded knowledge provided a secure basis for its answer; whereas in relation to lower resource jurisdictions, where it may not have embedded knowledge on which to depend, it would be more open to persuasion. Simplifying greatly, the former questions are plausibly closer to factual ones, where the model `knows' the answer, while the latter are closer to moral or policy questions, where evaluative judgment dominates. Existing work suggests models are more easily persuaded on subjective than objective questions~\citep{Zhao2025}. Conversely, if we observe higher persuadability in jurisdictions where models are assumed to have greater legal knowledge, this may indicate that the substantive legal content of arguments, as opposed to merely rhetorical style, is playing a significant role in persuasion.

\subsection{Argument Generation with LLM Advocates}

Given these summaries, we next generate a set of arguments on each side of each dispute. To do this, we use four different models as `Advocates': OpenAI GPT-4o, Google Gemini-3-pro-preview (4,096 thinking budget), OpenAI GPT-5.1 (low reasoning effort), and Anthropic Claude-sonnet-4.5 (8,192 thinking budget). We chose these models based on preliminary studies that identified an apparently significant difference in the persuasiveness of specific models. Our goal in selecting these models is not to identify the most effective advocate, but rather to ensure a range of advocates with varying abilities, to in turn map how far our Judge models are affected by the quality of the arguments presented to them.

We prompt each of our Advocate models to generate the most convincing arguments that it can for one or other party in each dispute. We use two different versions of each summary to prompt our Advocates. In the first version, we present only the three factual paragraphs from our summaries, omitting the summary of legal arguments actually advanced in the case, but including a short statement (also generated by o1) of the central legal question in the case. The template for the prompt to generate this argument is contained in Appendix~\ref{app:advocate_template_noarguments}. In the second version, we present the full summary, including both the facts and the legal arguments from the actual case. This prompt template is in Appendix~\ref{app:advocate_template_arguments}.

The rationale for this approach is that there are two distinct ways that we hypothesise one Advocate might be more effective than another. First, one Advocate might identify a novel point that the other did not address; or second, it might make the same point(s), but do so more convincingly, whether based on rhetorical style, logical structure or otherwise. In the former case, the novel point raised by the Advocate allows the Judge to consider a perspective that might not otherwise occur to them, plausibly improving the resulting decision. In the latter case, the more convincing presentation makes the Judge more likely to agree with the Advocate, but does not provide any additional information that might improve the decision.

By separately prompting our Advocates with and without the summary of arguments actually advanced, we get insight into which of these mechanisms dominates. We assume that the summaries that include the arguments actually advanced in the case will include the `best' arguments on both sides of the question: these cases have typically been litigated by leading professionals in their respective jurisdictions, and the judicial opinions will describe the arguments which successfully convinced the appellate judges on both sides of the split decision. By presenting our Advocate models with these argument summaries, we therefore provide `hints' about what are likely to be the best arguments to advance. If a Judge's preference for an Advocate reflects the content of their arguments, we would expect this additional context to improve the relative performance of the weaker Advocate, and hence reduce the differences between Advocate models. Conversely, if the Judge's preference is a function of the form in which an Advocate presented those arguments, the provision of additional context at the point of argument generation would make less difference. By comparing performance with or without this additional context, we shed light on which features are persuading our Judge models.

For each of the 15 scenarios, and each prompt version (with and without Advocate access to the original arguments), each model generates 5 arguments for each side of the dispute, for a total of 300 arguments per Advocate, or 1,200 arguments in total. We use this bank of arguments, of \emph{ex hypothesi} varying quality, to test the persuadability of our Judge models.

\subsection{Decision-Making with LLM Judges}

For each test, we randomly select a scenario, and within that scenario we randomly draw one argument for each of the parties, ensuring that the arguments are produced by different Advocate models. We then prompt each of the Judge models in turn with a prompt combining the facts, the statement of the central legal issue, and the two Advocate arguments (with the order of these randomised), and an instruction to decide the case as a court in the relevant jurisdiction would. Each randomly generated prompt is presented to each of the Judge models in turn, ensuring that all Judge models have received precisely the same stimuli, allowing us to directly compare results across Judge models. The template for generating the prompts for the Judge models can be seen in Appendix~\ref{app:judge_template}.

We test 20 models or model set-ups as Judges, which are listed in Table~\ref{tab:judge_models}. Whereas for our Advocate models, we sought to identify a number of models of differing abilities, for our Judge models we aim to test as wide a range as feasible of different models, on the basis that those seeking to put LLMs into production as decision-supports or decision-makers must choose from a panoply of available models. Three principal dimensions along which we sought to ensure variation in our sampling of Judge models included model size (we test both smaller and larger models within the same model families), model reasoning (we test analogous reasoning and non-reasoning models, or the same models with higher and lower reasoning settings/budgets), and model openness (we test both frontier closed-weight models, and also open-weight models from multiple providers). Varying model size and reasoning architecture reflects findings in the existing literature that these are significant in explaining persuadability. Including both closed and open-weights models reflects the fact that the former typically achieve higher performance on standard benchmarks, while users deploying legal decision tools in practice are likely to prefer the latter for \emph{inter alia} security reasons.

\begin{table}
\centering
\caption{20 Model configurations used as Judges. Each model is given a 3-letter code to indicate its type: L/S signifies a large or small model; R/N signifies a model with thinking/reasoning enabled or one with minimal or no reasoning; O/C signifies open or closed weights.} \label{tab:judge_models}
\centering

\footnotesize

\begin{tabular}{lllll}

\textbf{\#}   &  \textbf{Code} &   \textbf{Model}   &   \textbf{Thinking/Reasoning}  & \textbf{Version}     \\

\multicolumn{5}{c}{} \\
\cmidrule{1-5}
\multicolumn{5}{c}{\textbf{Large Closed Weights}}          \\

1 & LRC & \multirow[t]{2}{*}{Claude-sonnet-4.5}              & thinking budget=8192          & 20250929  \\
2 & LNC &                                                            & thinking budget=1024          & 20250929  \\

3 & LRC & \multirow[t]{2}{*}{Gemini-3-pro-preview}           & thinking budget=8192          &                   \\
4 & LNC &                                                            & thinking budget=1024          &                   \\

5 & LRC & \multirow[t]{2}{*}{GPT-5.1}                        & reasoning\_effort=medium      &                   \\
6 & LNC &                                                            & reasoning\_effort=low         &                   \\

\multicolumn{5}{c}{} \\

\cmidrule{1-5}

\multicolumn{5}{c}{\textbf{Large Open Weights}}          \\

7 & LRO & Deepseek-reasoner                          & reasoning model               & 3.1       \\

8 & LNO & Deepseek-chat                              & non-reasoning model           & 3.1       \\

9 & LRO & \multirow[t]{2}{*}{Qwen3-32B}              & thinking mode enabled         &                   \\
10 & LNO &                                          & thinking mode disabled        &                   \\

\multicolumn{5}{c}{} \\

\cmidrule{1-5}

\multicolumn{5}{c}{\textbf{Small Closed Weights}}          \\

11 & SRC & \multirow[t]{2}{*}{Claude-haiku-4.5}      & thinking budget=8192          & 20251001  \\
12 & SNC &                                          & thinking mode disabled         & 20251001  \\

13 & SRC & \multirow[t]{2}{*}{Gemini-2.5-flash-lite} & dynamic thinking enabled      &                   \\
14 & SNC &                                           & dynamic thinking disabled     &                   \\

15 & SRC & \multirow[t]{2}{*}{GPT-5-nano}            & reasoning\_effort=medium      &                   \\
16 & SNC &                                          & reasoning\_effort=minimal     &                   \\

\multicolumn{5}{c}{} \\

\cmidrule{1-5}

\multicolumn{5}{c}{\textbf{Small Open Weights}}          \\

17 & SRO & Magistral-Small-2506                      & thinking model                 &                   \\
18 & SNO & Mistral-Small-3.2-24B-Instruct-2506       & non-thinking model             &                   \\

19 & SRO & \multirow[t]{2}{*}{Qwen3-8B}              & thinking mode enabled          &                   \\
20 & SNO &                                          & thinking mode disabled         &                   \\

\end{tabular}
\end{table}

The prompts provided to the Judge models instructed that the last line should contain only the name of the party that it judged should prevail. Where this instruction was followed exactly, and the last line of the output matched a party to the case, this was taken as the outcome. In some cases, additional formatting or words were also included (e.g. ``Final answer: \textit{\{party name\}}'' or ``\textit{\{party name\}} should prevail''), but the preferred party was also clearly stated. This included minor errors in the names of parties that did not introduce any ambiguity as to who the preferred party was (e.g. ``The Secretary of \textit{the} State for the Home Department'' instead of ``The Secretary of State for the Home Department''). Occasionally a misinterpretation of the prompt resulted in a verbatim last line of ``\textit{\{party 1\}} or \textit{\{party 2\}}'', or additional output would appear at the end (including a legal disclaimer on rare occasions). In this situation, a manual inspection of the output was conducted to ascertain whether the name of the preferred party was unambiguously and explicitly stated towards the end of the output. Where the name was not explicit, we opted against performing manual judgments to infer a party name from the model output and instead discarded that trial. As this was the case for just 0.15\% of the trials (35 trials), it does not significantly affect the results. Another reason for this decision was that it was observed that occasionally the provided name appeared to be contrary to the reasoning process that preceded it. Thus including manual judgment would have required a re-examination or verification of all 24,000 trials. In addition to being overly burdensome, this would have introduced unnecessary and unhelpful subjectivity to the experiment. As a general observation, larger models tended to follow the output instructions more correctly than smaller ones, and all the discarded trials related to small models (in accordance with the categorisation in Table~\ref{tab:judge_models}).

\section{Metrics and Measuring Persuadability} \label{sec:metrics}

Our central concern is the extent to which a Judge model is influenced by the arguments presented on each side of a legal question. Given the varying persuasive capacities of our Advocate models, we adopt advocate identity as a proxy for argument quality, defining persuadability as the extent to which a Judge model's decision is affected by the identity of the Advocate models. If a Judge model was entirely non-persuadable (i.e., its decision was never affected by the arguments presented to them) then, given random assignment of Advocates to each side of each scenario, we would expect any given Advocate to be successful 50\% of the time (as nothing about the advocate would affect the outcome). Conversely, if we find that some Advocates are successful significantly more or less than 50\% of the time, this indicates that the Judge is more or less likely to accept that model's arguments, independent of the  merits of the case (as each model appears randomly on either side of each case), and is to that extent persuadable (and indeed persuaded).

So understood, persuadability is not simply a feature of a given Judge model. Rather, it is a function of the Judge model, given a specific pair of Advocate models. More formally, for any given pair of Advocate models, we define a Judge model's Pairwise Persuadability, $p_2$,  as:

\begin{equation}
    p_2 = \frac{|m_1 - m_2|}{2n}
\end{equation}

where $m_1$ and $m_2$ are the number of times that each Advocate model was successful against the other in our trial, and $n$ is the number of trials in which this specific pair of Advocate models was presented to the relevant Judge model. Pairwise persuadability thus takes a value between 0 and 0.5, representing the proportionate departure from the 50\% success rate that we would see if the Judge model was wholly non-persuadable. So, for example, if one model wins 75\% of the time, $p_2$ would be 0.25. If one model wins 100\% of the time, $p_2$ would be 0.5. And if both models were equally successful, $p_2$ would be 0 (i.e., the Judge model is not persuadable).

For a given population of Advocate models, we define the Judge model's Population Persuadability, $p_{pop}$, as the extent to which, across all model pairs, the model favours one model over the other. More formally:

\begin{equation}
    p_{pop} = sum(1 \rightarrow N) \frac{|m_1 - m_2|}{2n_{pop}}
\end{equation}

where $N$ is the set of all model pairings, and $n_{pop}$ is the total number of trials across all model pairings. As with Pairwise Persuadability, Population Persuadability takes a value between 0 and 0.5 representing the average departure from a 50/50 result, across all possible pairs given the population of Advocate models.

As will be clear from these definitions, judgments of persuadability are always context dependent, depending not only on the substantive questions being considered, but also on the range of Advocate models used. Our approach differs from several existing studies, which measure persuasion success by the difference between an agent's/model's agreement with a proposition before and after the persuasive interaction~\citep{Karande2024,Zhu2025,Zhao2025}. Given the importance of parties' arguments in legal settings, we do not consider that the pre-persuasion baseline is suitable in this context, while the availability of multiple Advocate models enables our alternative metric. 

One weakness of this metric is that, while it can identify that a Judge is likely to be persuaded by a stronger argument, it cannot distinguish between two ways that a Judge might appear non-persuadable. First, a Judge might appear non-persuadable because it relies on its own judgment and analysis, and is slow to depart from this in response to arguments from others, regardless of their quality. (This is what dialogical measures of persuadability measure.) Second, a Judge might appear non-persuadable on our metric because it is incapable of evaluating competing arguments, and instead, in the face of competing arguments, make decisions by randomising or relying on positional or other irrelevant biases. 

When evaluating models for deployment as decision assistants, the difference between these two forms of non-persuadability is critical. One may be a feature of a highly competent legal reasoner, while the other indicates a fundamental reasoning failure. Simply measuring trilateral persuadability would not allow us to distinguish between these two scenarios. To address this, we extend our core analysis through a parallel bilateral trial, using arguments from our main experiment, as well as trials varying party order to identify positional influence on results. As reported below, these supplementary analyses provide evidence that, in the case of at least some models, non-persuadability may indeed reflect a failure to distinguish among stronger and weaker arguments, while also providing clear evidence that the persuadability of models importantly has different features in trilateral settings, when compared with the bilateral dialogue more often tested in existing studies.

\section{Results} \label{sec:results}

This section begins with the presentation and analysis of the overall results of our experiments. This analysis then motivates more detailed investigations, including a breakdown by scenario (Section~\ref{sec:results:scenario}), a study on the effects of the order in which the parties' arguments are presented to the Judge models (Section~\ref{sec:results:order}), a contrast with a bilateral persuasion set-up (Section~\ref{sec:results:bilateral}), and analysis of the features of the arguments that influence persuadability (Section~\ref{sec:results:features}).

\subsection{Overall Results} \label{sec:results:overall}

Table~\ref{tab:persuadability_metrics} reports the Population Persuadability ($p_{pop}$) and Maximum Pairwise Persuadability ($p_2max$) of each model, in setups where Advocates were provided with the summary of facts and central legal issue from the original case, and also where they were given the summary of the arguments that were actually made in that case\footnote{Full results including confidence intervals and $p$~values for all Advocate pairs are available in our data repository.}. We report both $p_{pop}$ and $p_2max$ on the basis that the former measures the average persuadability across all pairings of our four Advocate models, while the latter measures persuadability across the pairing of the strongest and weakest Advocate models. As noted above, persuadability metrics will always be relative to the persuading agents. While neither metric is thus robust to variation in the Advocate population, $p_{pop}$ gives an overall sense of persuadability given a mixed population of Advocates, while $p_2max$ shows us what we might consider the ``worst case'', of a highly competent advocate and a significantly weaker one.
\begin{table}
\centering
    \caption{Maximum Pairwise Persuadability ($p_2max$) and Population Persuadability ($p_{pop}$) for all Judge models. All results significant ($p<0.05$). For $p_2max$ we use a binomial test for significance. For $p_{pop}$ we use a Chi Squared test to compare observed win rates of each advocate model with a 0.5 expected win rate under a null hypothesis of no persuasion.} \label{tab:persuadability_metrics}
    \centering
\begin{tabular}{clrrrr}
 & \multirow{2}{*}{\textbf{Judge Model}} &
 \multicolumn{2}{c}{\shortstack{\textbf{Without}\\\textbf{Arguments}}} &
 \multicolumn{2}{c}{\shortstack{\textbf{With}\\\textbf{Arguments}}}\\
 \cmidrule{3-6}

 & & \textbf{$p_2max$}& \textbf{$p_{pop}$}& \textbf{$p_2max$}&\textbf{$p_{pop}$} \\
 \cmidrule{1-6}
\multirow{4}{*}{\rotatebox{90}{\parbox[c]{1.4cm}{\centering Large\\Closed}}} 
& Claude-sonnet-4.5\_8k-thinking (LRC)             & 0.2328    & 0.1033    & 0.2500    & 0.1067    \\
& Claude-sonnet-4.5\_1k-thinking (LNC)             & 0.2931    & 0.1283    & 0.2759    & 0.1167    \\
& Gemini-3-pro-preview\_8k-thinking (LRC)          & 0.1724    & 0.1100    & 0.1638    & 0.1017    \\
& Gemini-3-pro-preview\_1k-thinking (LNC)          & 0.1552    & 0.1083    & 0.2328    & 0.1233    \\
& GPT-5.1\_medium-reasoning (LRC)                  & 0.2241    & 0.1083    & 0.1810    & 0.0950    \\
& GPT-5.1\_low-reasoning (LNC)                     & 0.2328    & 0.1217    & 0.2069    & 0.1133    \\
\cmidrule{1-6}
\multirow{4}{*}{\rotatebox{90}{\parbox{0.8cm}{\centering Large Open}}}
& Deepseek-reasoner (LRO)                          & 0.3707    & 0.1850    & 0.3276    & 0.1583    \\
& Deepseek-chat (LNO)                              & 0.3966    & 0.1850    & 0.4052    & 0.1983    \\
& Qwen3-32B\_thinking (LRO)                        & 0.3190    & 0.1950    & 0.2586    & 0.1683    \\
& Qwen3-32B\_nothinking (LNO)                      & 0.2845    & 0.1517    & 0.2500    & 0.1433    \\
\cmidrule{1-6}
\multirow{4}{*}{\rotatebox{90}{\parbox[c]{1.4cm}{\centering Small\\Closed}}}
& Claude-haiku-4.5\_8k-thinking (SRC)              & 0.3190    & 0.1700    & 0.3276    & 0.1567    \\
& Claude-haiku-4.5\_nothinking (SNC)               & 0.2586    & 0.1250    & 0.3362    & 0.1533    \\
& Gemini-2.5-flash-lite\_reasoning (SRC)           & 0.3522    & 0.2008    & 0.3435    & 0.1857    \\
& Gemini-2.5-flash-lite\_nonreasoning (SNC)        & 0.3000    & 0.1454    & 0.2652    & 0.1341    \\
& GPT-5-nano\_medium-reasoning (SRC)               & 0.2155    & 0.1350    & 0.2155    & 0.1350    \\
& GPT-5-nano\_minimal-reasoning (SNC)              & 0.1311    & 0.0893    & 0.1667    & 0.0800    \\
\cmidrule{1-6}
\multirow{4}{*}{\rotatebox{90}{\parbox{0.8cm}{\centering Small Open}}}
& Magistral-Small-2506 (SRO)                       & 0.2155    & 0.1093    & 0.1810    & 0.1003    \\
& Mistral-Small-3.2-24B-Instruct-2506 (SNO)        & 0.2157    & 0.1171    & 0.2345    & 0.0993    \\
& Qwen3-8B\_thinking (SRO)                          & 0.2051    & 0.1183    & 0.2155    & 0.1250    \\
& Qwen3-8B\_nothinking (SNO)                       & 0.2130    & 0.1244    & 0.2759    & 0.1333    \\
    \end{tabular}
\end{table}

We first observe that, across all models, both $p_{pop}$ and $p_2max$ are statistically significant. All of our models are, to some extent, persuadable. $p_{pop}$ ranges from 0.08 up to 0.2008, while $p_2max$ runs from 0.1311 to 0.4052. This means that, across our full range of models, the identity of the Advocate model (and hence the quality of the argument presented) has an average effect of between 8\% and 21\%, implying that stronger Advocate models typically win between 58\% and 71\% of the time. Further, as between the strongest and weakest Advocate models, depending on the Judge model, those win rates range from 63\% to over 90\%. We therefore conclude that all our Judge models are to some extent persuadable. Further, we note that in all cases, these persuasive effects are statistically significant.

\begin{figure*}
    \centering
    \includegraphics[width=\linewidth]{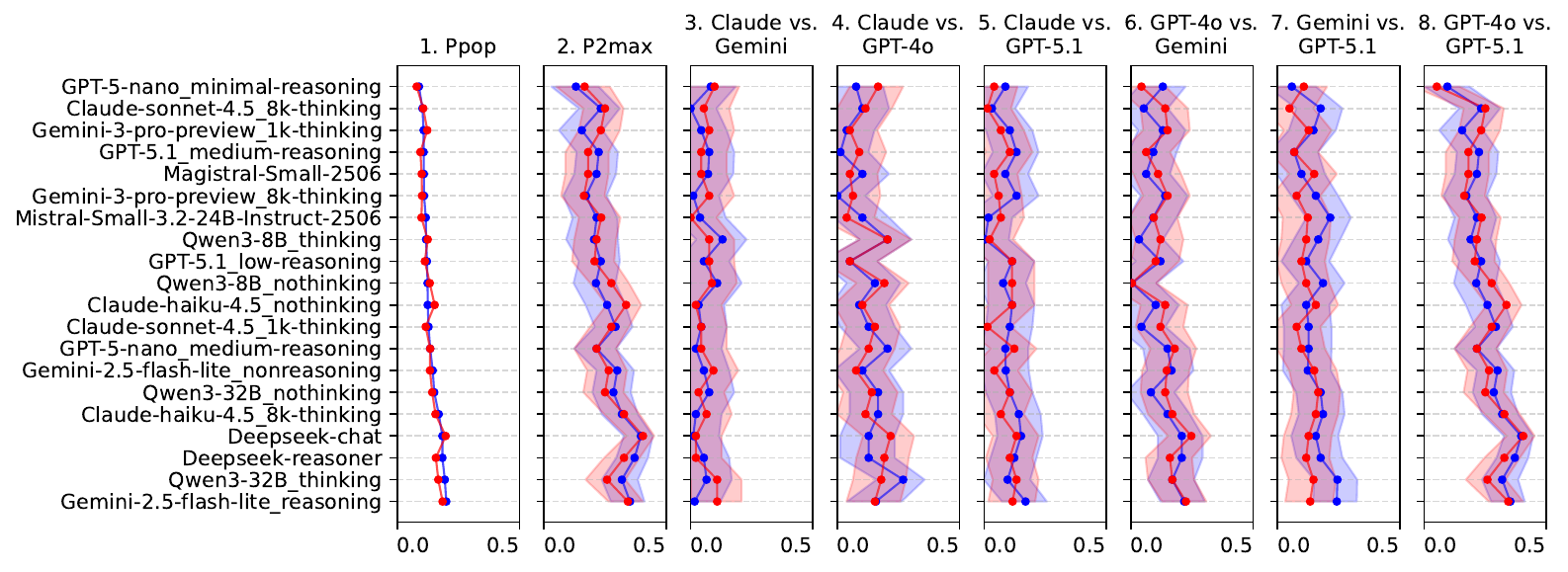}
    \caption{Population persuadability ($p_{pop}$), maximum pairwise persuadability ($p_2max$) and pairwise persuadability ($p_2$) by Advocate pair for each Judge model. 
    Blue: Without Original Arguments; Red: With Original Arguments} 
    \label{fig:persuadibility}
\end{figure*}

As is clear from Table~\ref{tab:persuadability_metrics}, the extent of persuadability varies substantially across Judge models. Pairwise persuadability ($p_2$) also varies depending on the Advocate models involved. Again, this is to be expected: our Advocate models were selected to ensure a range of persuasiveness, so it is unsurprising to find larger values for $p_2$ where a stronger Advocate is paired with a weaker one than for two Advocates of similar strength. Figure~\ref{fig:persuadibility} presents the pairwise persuadability ($p_2$) (with 95\% confidence intervals) for each Judge model across each Advocate pairing, together with the maximum pairwise persuadability ($p_2max$) and population persuadability ($p_{pop}$). From this we readily observe that $p_2$ is typically lowest for Claude vs Gemini: these are two similarly persuasive models. By contrast, $p_2$ is in most cases highest for GPT-4o vs GPT-5.1, where there is the greatest performance gap between Advocates.

We also observe some support for the hypothesis that larger models are less persuadable than smaller models. However, the evidence here is mixed. Amongst our closed models, where we test pairs of larger and smaller models from the same model families, we see, in most cases, the larger/full model is less persuadable than the smaller/lite model. The primary exception is GPT-5\_nano\_minimal-reasoning, which has the lowest $p_{pop}$ of any model in our trial, in both the With and Without Original Arguments conditions. The other exception is the marginally higher $p_{pop}$ for Claude-sonnet with a low thinking budget, compared to Claude-haiku with thinking disabled. Amongst our open models, the Qwen3-8B models appear less persuadable than Qwen3-32B, across both thinking and non-thinking settings. Mistral and Magistral are two of the smaller models in our trial, but also amongst the least persuadable.

Similarly, we observe some support for the hypothesis that models with reasoning architecture are less persuadable than those without, particularly for larger model sizes. Amongst our large closed models, a higher reasoning setting / thinking budget corresponds to lower persuadability in five of six cases (the exception is Gemini-3-pro in the setup without original arguments). For our smaller closed models and a number of our open models (Qwen3-32B, Mistral/Magistral) this relationship is reversed, with higher reasoning variants appearing more persuadable on our metric. However, in only one case (GPT-5-nano with original arguments) is this difference statistically significant at the population level.\footnote{Chi Squared test, $p<0.05$. Full significance tests are included in the data repository.}

We hypothesise that these unexpected results in relation to both model size and reasoning architecture reflect the different form that persuadability takes in our experiments. Bilateral persuasion scenarios test a model's pre-interaction response, and then present it with arguments tending in one direction only, measuring how far its response changes given this persuasive treatment. In that setup, a model does not need to be capable of evaluating the arguments presented to it in order to be persuaded by them. By contrast, in our setup models are presented with competing arguments. In order to be persuaded by the stronger argument, a model must be capable of evaluating those two competing arguments to identify which is in fact stronger. For a model that is not capable of evaluating arguments, we would expect the two competing arguments to be equally convincing, and hence lower overall persuasion. In Section~\ref{sec:results:bilateral} we present results of a bilateral trial that lends credence to this hypothesis.

\subsection{Results by Scenario} \label{sec:results:scenario}

As a first test of the hypothesis that low persuadability may sometimes reflect a failure to distinguish strong from weak arguments, we examine Judge model results at the level of individual scenarios. We reason as follows: if a model appears less persuadable because it is forming its own view in relation to a given scenario, then we would expect to find a statistically significant preference for one outcome rather than the other in that scenario. By contrast, if a model shows low persuadability in our tests, but also has no statistically significant preference over outcome in a given scenario, this suggests that the model is to a large extent randomising in its responses. Further, because these scenarios present hard questions, on which competent experts disagree, we would expect that a model that was forming and holding to its own view would have a preference that varies between the first and second parties across the fifteen scenarios. By contrast, if a model consistently favoured either the first or the second party, this would suggest that positional bias was significantly affecting its decisions. Finally, in our setup, the side that won in the real case is always presented as the first party.  We might thus be more inclined to interpret a consistent preference for the first party as reflecting a reasoned preference for that party's position, while a consistent preference for the second party (the losing party in the real case) more strongly suggests that positional bias is involved.

Table~\ref{tab:case-preference} presents, for each Judge model, first, the proportion of scenarios where the relevant model has a significant preference for one or other party, and second, the proportion of those scenarios where that significant preference favoured the first party. 

\begin{table}
 \caption{Proportion of 15 scenarios in which each Judge has a statistically significant preference for one party over the other and, of scenarios where a preference is significant, proportion that favour Party 1.}     \label{tab:case-preference} 
    \centering
    \begin{tabular}{lrrrr}
    
 \multirow{2}{*}{\textbf{Judge Model}} &
 \multicolumn{2}{c}{\shortstack{\textbf{Without}\\\textbf{Arguments}}} &
 \multicolumn{2}{c}{\shortstack{\textbf{With}\\\textbf{Arguments}}}\\
 
 \cmidrule{2-5}
    &
    \multicolumn{1}{p{1.37cm}}{\centering Significant Preference} &  
    \multicolumn{1}{p{1.37cm}}{\centering Party 1 Preference} &
    \multicolumn{1}{p{1.37cm}}{\centering Significant Preference} &
    \multicolumn{1}{p{1.37cm}}{\centering Party 1 Preference}   \\

\cmidrule{1-5}

Qwen3-32B\_thinking (LRO)                   & 0.47  & 0.29  & 0.67  &   0.40    \\
Deepseek-reasoner (LRO)                     & 0.53  & 0.50  & 0.53  &   0.63    \\
Gemini-2.5-flash-lite\_reasoning (SRC)      & 0.53  & 0.50  & 0.60  &   0.44    \\
GPT-5-nano\_minimal-reasoning (SNC)         & 0.53  & 0.38  & 0.40  &   0.17    \\
Magistral-Small-2506 (SRO)                  & 0.53  & 0.00  & 0.73  &   0.27    \\
GPT-5-nano\_medium-reasoning (SRC)          & 0.60  & 0.22  & 0.73  &   0.45    \\
Deepseek-chat (LNO)                         & 0.67  & 0.40  & 0.53  &   0.25    \\
Gemini-2.5-flash-lite\_nonreasoning (SNC)   & 0.73  & 0.55  & 0.53  &   0.63    \\
Gemini-3-pro-preview\_1k-thinking (LNC)     & 0.73  & 0.36  & 0.80  &   0.50    \\
GPT-5.1\_medium-reasoning (LRC)             & 0.73  & 0.45  & 0.80  &   0.50    \\
Mistral-Small-3.2-24B-Instruct-2506 (SNO)   & 0.73  & 0.27  & 0.67  &   0.20    \\
Qwen3-8B\_nothinking (SNO)                  & 0.73  & 0.64  & 0.73  &   0.55    \\
Qwen3-8B\_thinking (SRO)                    & 0.73  & 0.82  & 0.60  &   0.78    \\
Claude-haiku-4.5\_8k-thinking (SRC)         & 0.80  & 0.33  & 0.60  &   0.22    \\
Gemini-3-pro-preview\_8k-thinking (LRC)     & 0.80  & 0.50  & 0.73  &   0.55    \\
Qwen3-32B\_nothinking (LNO)                 & 0.80  & 0.50  & 0.87  &   0.46    \\
Claude-sonnet-4.5\_1k-thinking (LNC)        & 0.87  & 0.46  & 0.73  &   0.36    \\
Claude-sonnet-4.5\_8k-thinking (LRC)        & 0.87  & 0.46  & 0.80  &   0.42    \\
GPT-5.1\_low-reasoning (LNC)                & 0.87  & 0.38  & 0.80  &   0.42    \\
Claude-haiku-4.5\_nothinking (SNC)          & 0.93  & 0.36  & 0.73  &   0.36    \\
    \end{tabular}

\end{table}

The larger closed models appear somewhat more likely to have a statistically significant preference for one or other party, but the relation here is not strong, and there is no obvious pattern in relation to which models have preferences that vary between positions, versus those that have an apparent positional bias. We do, however, note that GPT-5-nano\_minimal-reasoning, which was an outlier in relation to persuadability, has a relatively low (0.5333 and 0.4) proportion of scenarios in which it has a statistically significant preference for one party, and a low proportion of these (0.375 and 0.17) in which it favours Party 1, which may support our hypothesis that its low persuadability indicates randomising and/or positional bias rather than adherence to a fixed preference. The Qwen3-8B models, which also appeared less persuadable than the larger Qwen3-32B models, have a relatively high (0.6-0.733) proportion of scenarios in which they have a significant preference, and of these, they either have a roughly equal mix of preferences, or a tendency to prefer the first party over the second party. This is less readily explained in terms of randomising. It might be ascribed to positional bias in favour of the first set of arguments, but the better explanation is probably that this model is indeed less persuadable than its larger counterpart.

\subsection{Effects of Party Order} \label{sec:results:order}

To further explore the role of positional bias \citep{Shi2025}, and more generally how far party order is impacting our results, we ran a smaller supplemental test, re-prompting our models with all of the trials that, in the original experiment, paired GPT-4o and GPT-5.1 as Advocates, but reversing the party and argument order from that in our original experiment. This allows us to directly test for positional bias. We use the GPT-4o/GPT-5.1 pairing as this has the largest advocate effect, and therefore presents the hardest test for positional bias. If we observe positional effects across this pairing we can be confident we would find them with the other pairings also. As reported in Table~\ref{tab:party_order} (second column), we find significant positional bias in eight of our Judge models. Gemini-3-pro\_thinking, Qwen3-8B\_thinking and Deepseek-reasoner each show a preference for the first party, while Mistral, Magistral, Claude-haiku\_thinking, Claude-haiku\_nothinking and Qwen3-32B\_thinking each show a preference for the second party.

However, going beyond consistent positional bias, reversing party order has a substantial impact on specific model outputs. As reported in Table~\ref{tab:party_order} (first column)\footnote{For these trials, Gemini-3.1-pro-preview was used in place of Gemini-3-pro-preview as the latter was decommissioned before this follow-on experiment was conducted. For a valid comparison, the original prompts with the original party ordering were also re-run with Gemini-3.1-pro-preview.}, in a substantial proportion of trials (ranging from 0.112 up to 0.578), reversing the order of parties and arguments resulted in the model switching its conclusion. This means that, in between 11\% and 58\% of trials, which party wins depends on the order in which the arguments are presented. In some cases, this reflected a consistent positional bias, but in others it simply highlights the instability of model outputs given minor and legally non-significant changes in inputs. To confirm that the effect here is a function of changing party and argument order, rather than simply instability in model outputs, we contrast these results with the observed consistency of our models when prompted with no arguments. Table~\ref{tab:party_order} (last column) reports the proportion of trials where the model returned its most commonly returned response for the chosen scenario, without any arguments being provided. As will be clear, in almost all cases the variance resulting from reversing argument order is substantially greater than that which emerges in the no-arguments setup. We thus conclude that varying the order of arguments has significant impacts on model outputs, and that these impacts go beyond any consistent positional bias exhibited by our Judge models.

\begin{table}
\centering
\caption{Effects of Party and Argument Order on Judge Model Outputs
(i) Proportion of Trials where Reversing Party and Argument Order Changes Result (ii) Proportion of Trials Where Model Chooses the First Party presented; (iii) binomial $p$~value for positional bias (Significant ($p<0.05$) in bold); (iv) Proportion of Trials in No Arguments Setup in which Model Chooses More Common Outcome}
\label{tab:party_order}

\begin{tabular}{l c c c c}
 \textbf{Judge Model} & \shortstack{\textbf{Result} \\\textbf{reversed}} & \shortstack{\textbf{Party 1}\\\textbf{wins}} & \shortstack{\textbf{$p$~value}\\~} & \shortstack{\textbf{Pre-persuasion}\\\textbf{consistency}} \\
 \cmidrule{1-5}
Claude-sonnet-4.5\_8k-thinking       (LRC) & 0.181 & 0.435 & 0.057 & 0.970 \\
Claude-sonnet-4.5\_1k-thinking       (LNC) & 0.164 & 0.453 & 0.168 & 0.950 \\
Gemini-3.1-pro-preview\_8k-thinking  (LRC) & 0.112 & 0.513 & 0.743 & 0.920 \\
Gemini-3.1-pro-preview\_1k-thinking  (LNC) & 0.112 & 0.513 & 0.743 & 0.890 \\
GPT-5.1\_medium-reasoning            (LRC) & 0.172 & 0.509 & 0.948 & 0.930 \\
GPT-5.1\_low-reasoning               (LNC) & 0.138 & 0.526 & 0.470 & 0.890 \\
Deepseek-reasoner                    (LRO) & 0.241 & 0.569 & \textbf{0.042} & 0.880 \\
Deepseek-chat                        (LNO) & 0.164 & 0.530 & 0.393 & 0.920 \\
Qwen3-32B\_thinking                  (LRO) & 0.259 & 0.405 & \textbf{0.005} & 0.940 \\
Qwen3-32B\_nothinking                (LNO) & 0.198 & 0.453 & 0.168 & 0.880 \\
Claude-haiku-4.5\_8k-thinking        (SRC) & 0.216 & 0.409 & \textbf{0.007} & 0.890 \\
Claude-haiku-4.5\_nothinking         (SNC) & 0.233 & 0.409 & \textbf{0.007} & 0.900 \\
Gemini-2.5-flash-lite\_reasoning     (SRC) & 0.272 & 0.496 & 0.948 & 0.850 \\
Gemini-2.5-flash-lite\_nonreasoning  (SNC) & 0.261 & 0.552 & 0.131 & 0.870 \\
GPT-5-nano\_medium-reasoning         (SRC) & 0.310 & 0.543 & 0.212 & 0.860 \\
GPT-5-nano\_minimal-reasoning        (SNC) & 0.578 & 0.513 & 0.743 & 0.720 \\
Magistral-Small-2506                 (SRO) & 0.379 & 0.405 & \textbf{0.005} & 0.710 \\
Mistral-Small-3.2-24B-Instruct-2506  (SNO) & 0.354 & 0.353 & \textbf{0.000} & 0.880 \\
Qwen3-8B\_thinking                   (SRO) & 0.310 & 0.578 & \textbf{0.021} & 0.860 \\
Qwen3-8B\_nothinking                 (SNO) & 0.284 & 0.565 & 0.057 & 0.830 \\

\end{tabular}
\end{table}

\subsection{Contrasting Bilateral and Trilateral Persuasion Settings} \label{sec:results:bilateral}

As noted above in Section~\ref{sec:metrics}, we hypothesised that model persuadability would manifest differently in trilateral, compared with bilateral, settings. In Section~\ref{sec:results:overall} we noted some features of our trilateral results that appear to support this conclusion. To further test this hypothesis, we ran a set of supplemental trials, in which we expose our Judge models to bilateral persuasion, but using precisely the same scenarios and arguments as our principal, trilateral trials.

To do this we first prompt our models to decide a case without the benefit of  arguments for either party, presenting only the facts and the key legal issue (i.e. the same information provided to our advocates). This establishes a pre-persuasion baseline. We then pass two further prompts. In the first, we prompt our model with the facts, the key legal issue, and an argument for the opposite conclusion to that reached without any arguments, but without including any reference to the model's original decision. In the second, we prompt the model with a dialogue, including the original prompt without arguments, the model's response including its initial decision in the case, and a further user response setting out the argument for the alternative position. We use only arguments generated by GPT-5.1 for this trial, on the basis that these were the most persuasive in our trilateral trial. Using these two different approaches reflects a hypothesis that including the model's initial response will induce an anchor effect, making the model less persuadable than if we provide the one-sided argument without the initial response. In line with existing works we measure bilateral persuadability by comparing the pre-persuasion baseline with our two subsequent trials. 

Our results are set out in Table~\ref{tab:bilateral}.

\begin{table}
\centering

\caption{Persuadability Results in Bilateral Persuasion Setting. Proportion of trials in which model changed response after persuasion. \textit{AO}: where model was prompted with argument only; \textit{RA} where model was prompted with pre-persuasion response plus argument; \textit{Both} where model changed position in both set-ups.}
\label{tab:bilateral}

\begin{tabular}{l l l l}
\textbf{Judge Model} & \textbf{AO} & \textbf{RA} & \textbf{Both} \\
\cmidrule{1-4}
Claude-sonnet-4.5\_8k-thinking (LRC)     & 0.11 & 0.55 & 0.09 \\
Claude-sonnet-4.5\_1k-thinking (LNC)     & 0.18 & 0.66 & 0.18 \\
Gemini-3.1-pro-preview\_8k-thinking (LRC) & 0.07 & 0.10 & 0.06 \\
Gemini-3.1-pro-preview\_1k-thinking (LNC) & 0.11 & 0.11 & 0.07 \\
GPT-5.1\_medium-reasoning (LRC)          & 0.17 & 0.11 & 0.04 \\
GPT-5.1\_low-reasoning (LNC)             & 0.19 & 0.06 & 0.03 \\
Deepseek-reasoner (LRO)                  & 0.74 & 0.46 & 0.39 \\
Deepseek-chat (LNO)                      & 0.76 & 0.59 & 0.51 \\
Qwen3-32B\_thinking (LRO)                & 0.93 & 0.69 & 0.65 \\
Qwen3-32B\_nothinking (LNO)              & 0.81 & 0.65 & 0.57 \\
Claude-haiku-4.5\_8k-thinking (SRC)      & 0.44 & 0.75 & 0.38 \\
Claude-haiku-4.5\_nothinking (SNC)       & 0.29 & 0.88 & 0.28 \\
Gemini-2.5-flash-lite\_reasoning (SRC)   & 0.82 & 0.63 & 0.54 \\
Gemini-2.5-flash-lite\_nonreasoning (SNC)& 0.77 & 0.85 & 0.70 \\
GPT-5-nano\_medium-reasoning (SRC)       & 0.70 & 0.53 & 0.41 \\
GPT-5-nano\_minimal-reasoning (SNC)      & 0.82 & 0.77 & 0.64 \\
Magistral-Small-2506 (SRO)               & 0.86 & 0.92 & 0.80 \\
Mistral-Small-3.2-24B-Instruct-2506 (SNO)& 0.75 & 0.90 & 0.69 \\
Qwen3-8B\_thinking (SRO)                  & 0.72 & 0.78 & 0.57 \\
Qwen3-8B\_nothinking (SNO)               & 0.81 & 0.84 & 0.70 \\

\end{tabular}
\end{table}

We observe a number of important features. First, those smaller models that appeared less persuadable on our trilateral trial (GPT-5-nano\_minimal, Mistral, Magistral) now appear amongst the most persuadable. In our trilateral trial, the smaller (8B) Qwen models were less persuadable than their larger (32B) counterparts. In a bilateral setting, they remain less persuadable in our first trial (without the initial judge model output) but more persuadable in the second (with the initial output included). Overall, our bilateral results more closely reflect existing research identifying an inverse relationship between model size and persuadability. We continue to see some suggestive evidence that reasoning models are less persuadable than non-reasoning models, but the differences here are small and inconsistent, with reasoning models appearing less persuadable than the comparable non/low reasoning model in 19 of 30 pairings in Table~\ref{tab:bilateral}. Second, the hypothesised anchor effect of prompting a model with its own initial output is observed in the four GPT models, and in our four large open models. (Albeit these models' persuadability is low in both set-ups, so it is hard to make confident judgments.) However, the effect is smaller or reversed in the case of the four Gemini models and the small open-source models. Further, there appears to be a substantial inverse anchor effect in the case of the four Claude models, which are much more likely (31\% to 59\%) to change their position where they are presented with their own prior conclusion than when they are simply given an argument for the opposite position. This last result highlights the continued problems of sycophancy arising from the post-training of LLMs for helpfulness, as studied by~\cite{Fanous2025}. This is largely neutralised in a trilateral setting where there is not one user position with which the model can conform. More generally, the contrast between our bilateral and trilateral results confirms the importance of studying trilateral persuasion as a distinct phenomenon, in parallel with bilateral persuasion.

These results also support our hypothesis relating to trilateral versus bilateral persuasion, as outlined in Section~\ref{sec:results:overall}. The fact that the larger models tend to be far less persuadable than smaller ones in the bilateral context suggests that their own judgment and analysis is an important reason for their similar lack of trilateral persuadability. In contrast, the smallest models that were surprisingly non-persuadable in the trilateral setting are seen to be the most persuadable in our bilateral trials, suggesting that lower trilateral persuadability reflects a lesser ability to distinguish between competing arguments, thereby resulting in more random behaviour.

\subsection{Which Features of Arguments Persuade Judge Models?} \label{sec:results:features}

Having established the significant persuadability of all of our Judge models, we apply a number of further analyses to investigate argument features that may influence persuadability. This includes a study on whether the legal content of an argument is more or less significant than the rhetorical form it is presented in (Section~\ref{sec:results:content-vs-form}), an analysis of argument strength according to the viewpoint of qualified human experts (Section~\ref{sec:results:human}), and an investigation as to whether the quantity, type and diversity of the arguments advanced has an influence on persuadability (Section~\ref{sec:results:quantity-type-diversity}).

\subsubsection{Legal Content vs Rhetorical Form} \label{sec:results:content-vs-form}
We use two strategies to identify whether / how far persuadability is a function of Advocate models presenting novel arguments that the judge model might not have otherwise considered, and hence improving the quality of the decision, versus reflecting the greater fluency or rhetorical powers of the relevant models.

First, as outlined in Section~\ref{sec:results:overall} above, we use two different setups to prompt our Advocates. In one of these we provide only the facts, and in the other we present also a summary of the arguments actually made in the case. We observe (see Table~\ref{tab:persuadability_metrics}) that in 14 of 20 model setups, population persuadability is lower where Advocates were prompted with the arguments from the original case, suggesting that the content, as opposed to the form, of arguments is playing some role in persuasion. However, the difference is generally small, and in no case statistically significant (Chi Squared test, $p$<0.05). Disaggregating results by Advocate pair (see Figure~\ref{fig:persuadibility}) further illustrates that, to the extent there is a relationship here, it is a weak one.

To further test whether providing summaries of the arguments in the original case made a weaker Advocate more persuasive, we ran additional head-to-head trials involving the same Advocate model arguing against itself, with one version prompted with the arguments from the original case, and the other without. We thus directly test whether, given the same Advocate model, providing the arguments from the original case increases the likelihood of success. This ensures that the availability or otherwise of the original arguments are the only variable. We trial this setup with two Advocates, GPT-4o and GPT-5.1, and a selection of four Judge models. Results are presented in Table~\ref{tab:head-to-head}.

\begin{table}[b]
\caption{Head-to-head trials of Advocate models prompted with or without summaries of arguments from original case. Win rate of Advocate prompted with arguments. Binomial $p$~values.}
    
    \centering
    \begin{tabular}{lrrrr}
\multirow{2}{*}{\textbf{Judge Model}} &
\multicolumn{2}{c}{\textbf{GPT-4o}} &
\multicolumn{2}{c}{\textbf{GPT-5.1}} \\

\cmidrule{2-5}
                                    & Win rate  & $p$~value & Win rate  & $p$~value \\

\cmidrule{1-5}

Claude-sonnet-4.5\_8k (LRC)              & 0.5567    & 0.3099    & 0.5243    & 0.6937    \\
Deepseek-chat (LNO)                      & 0.6804    & 0.0005    & 0.5922    & 0.0756    \\
Gemini-2.5-flash-lite\_reasoning (SRC)   & 0.6042    & 0.0519    & 0.5098    & 0.9212    \\
GPT-5.1\_medium-reasoning (LRC)          & 0.5876    & 0.1038    & 0.5437    & 0.4307    
    \end{tabular}
    \label{tab:head-to-head}
\end{table}
Across eight head-to-head trials (pairing two Advocate models across four Judges), the model prompted with the arguments from the original case won more frequently than the model prompted without the argument summary in every trial. However, the treatment effects are small, with only one case (GPT-4o as Advocate, Deepseek-chat as Judge) reaching the threshold of significance ($p<0.05$). If we aggregate the overall outcome of each of the eight trials, applying a binomial test to this 8-0 result gives a $p$~value of 0.0039. i.e. it seems clear that, while the effect size is small, the provision of the original arguments when prompting our Advocates does affect the overall persuasive effect. We also observe that, across all Judges, the treatment effect appears larger in the case of GPT-4o, suggesting that providing the original argument summary is indeed helping the weaker models perform better, in turn implying that the substantive legal content of the argument, as opposed to mere form, plays at least some role. (Combining results by advocate across the four judges reinforces this: results for GPT-4o appear significant ($p<0.05$) in this setup, while those for GPT-5.1 do not.) 

Second, we disaggregate results by jurisdiction (US / England and Wales / Ireland) on the assumption that models have relatively greater knowledge of US law, and relatively lesser knowledge of Irish law. If we observe higher persuadability in jurisdictions where models are assumed to have greater knowledge, this suggests that persuasion is driven at least in part by the quality of substantive legal arguments, as opposed to rhetorical skill or form. A null result here would be less informative, as there may be different explanations of why persuasion might not vary across jurisdictions. Results of this trial are presented in Table~\ref{tab:jurisdictions} below. 

\begin{sidewaystable}[p]
    \caption{Population persuadability ($p_{pop}$) disaggregated by jurisdiction: United States (US$p_{pop}$), England and Wales (E\&W$p_{pop}$), Ireland (IRL$p_{pop}$). Chi squared $p$~values based on overall Advocate win rates for each Judge model.}
    \centering
    \begin{tabular}{lrrrrrrrr}
\multirow{2}{*}{\textbf{Judge Model}}    & \multicolumn{4}{c}{\textbf{Without Original Arguments}}& \multicolumn{4}{c}{\textbf{With Original Arguments}}\\
\cmidrule{2-9}
                                &  US$p_{pop}$& E\&W$p_{pop}$& IRL$p_{pop}$& $p$~value & US$p_{pop}$& E\&W$p_{pop}$& IRL$p_{pop}$& $p$~value \\
\cmidrule{1-9}
         
Claude-sonnet-4.5\_8k-thinking (LRC)         & 0.1968    & 0.0901    & 0.0684    & 0.0505    & 0.1755    & 0.1171    & 0.0632    & 0.0283    \\
Claude-sonnet-4.5\_1k-thinking (LNC)         & 0.2287    & 0.0856    & 0.0842    & 0.0122    & 0.2021    & 0.1396    & 0.0684    & 0.0075    \\
Gemini-3-pro-preview\_8k-thinking (LRC)      & 0.1915    & 0.1126    & 0.1053    & 0.2015    & 0.1862    & 0.0901    & 0.0737    & 0.2022    \\
Gemini-3-pro-preview\_1k-thinking (LNC)      & 0.1702    & 0.0946    & 0.0895    & 0.7092    & 0.1755    & 0.1081    & 0.1053    & 0.6658    \\
GPT-5.1\_medium-reasoning (LRC)              & 0.1755    & 0.0946    & 0.1526    & 0.2735    & 0.1330    & 0.0676    & 0.1158    & 0.7458    \\
GPT-5.1\_low-reasoning (LNC)                 & 0.1755    & 0.0991    & 0.1105    & 0.6038    & 0.1596    & 0.0811    & 0.1053    & 0.6049    \\
Deepseek-reasoner (LRO)                      & 0.1649    & 0.2162    & 0.1737    & 0.7160    & 0.1170    & 0.2072    & 0.1579    & 0.5470    \\
Deepseek-chat (LNO)                          & 0.2021    & 0.2568    & 0.1368    & 0.0143    & 0.1862    & 0.3063    & 0.1263    & 0.0120    \\
Qwen3-32B\_nothinking (LNO)                  & 0.1862    & 0.1757    & 0.1053    & 0.1589    & 0.1915    & 0.1441    & 0.1158    & 0.3347    \\
Qwen3-32B\_thinking (LRO)                    & 0.2606    & 0.2387    & 0.0895    & 0.0004    & 0.2979    & 0.1802    & 0.0632    & 0.0000    \\
Claude-haiku-4.5\_8k-thinking (SRC)          & 0.2553    & 0.2117    & 0.0789    & 0.0003    & 0.2447    & 0.1892    & 0.0895    & 0.0006    \\
Claude-haiku-4.5\_nothinking (SNC)           & 0.1649    & 0.1982    & 0.0526    & 0.0010    & 0.1862    & 0.2297    & 0.0842    & 0.0041    \\
Gemini-2.5-flash-lite\_reasoning (SRC)       & 0.1915    & 0.1941    & 0.2181    & 0.8512    & 0.2151    & 0.1804    & 0.1632    & 0.6193    \\
Gemini-2.5-flash-lite\_nonreasoning (SNC)    & 0.2166    & 0.1577    & 0.0645    & 0.0228    & 0.1237    & 0.1728    & 0.1000    & 0.7143    \\
GPT-5-nano\_medium-reasoning (SRC)           & 0.1011    & 0.2027    & 0.1316    & 0.1907    & 0.0745    & 0.1757    & 0.1684    & 0.6046    \\
GPT-5-nano\_minimal-reasoning (SNC)          & 0.1223    & 0.1126    & 0.0873    & 0.0585    & 0.1117    & 0.0721    & 0.0895    & 0.9974    \\
Magistral-Small-2506 (SRO)                   & 0.1436    & 0.1471    & 0.1053    & 0.0733    & 0.1543    & 0.0973    & 0.0926    & 0.2348    \\
Mistral-Small-3.2-24B-Instruct-2506 (SNO)    & 0.1596    & 0.1199    & 0.0926    & 0.1010    & 0.1471    & 0.1109    & 0.0591    & 0.6399    \\
Qwen3-8B\_thinking (SRO)                     & 0.1543    & 0.1532    & 0.0947    & 0.0244    & 0.1223    & 0.1712    & 0.0947    & 0.4561    \\
Qwen3-8B\_nothinking (SNO)                   & 0.1471    & 0.1216    & 0.1421    & 0.4650    & 0.1330    & 0.1261    & 0.1737    & 0.6215

\end{tabular}

    \label{tab:jurisdictions}
\end{sidewaystable}

For 13 of our 40 Judge model set-ups, there is statistically significant variation across jurisdictions. Further, as reported in Table~\ref{tab:jurisdictions}, the variation in most cases is in the direction which we would expect if substantive legal content explained at least part of the observed persuadability. i.e. $p_{pop}$ is in most cases lower for our Irish scenarios than for our E\&W scenarios, which are in turn lower than our US scenarios. In 28 (of 40) cases, E\&W$p_{pop}$ is lower than US$p_{pop}$, and in 32 cases IRL$p_{pop}$ is lower than E\&W$p_{pop}$. In 34 cases, IRL$p_{pop}$ is lower than US$p_{pop}$. We thus again find suggestive evidence that the quality of substantive legal arguments, as opposed to mere rhetorical skill, plays a role in model persuadability. However, the fact that this effect is significant in less than one third of our Judge model set-ups, representing only eight of our Judge models (all four Claude models, Gemini-2.5-flash-lite\_non-reasoning, Deepseek-chat and the two Qwen reasoning models) prevents us drawing any strong or general conclusions from this test.

\subsubsection{Human Argument Quality Judgment and Model Persuasion} \label{sec:results:human} 

We next ask whether Advocate persuasiveness vis-\`a-vis our Judge models tracks human expert judgments of argument quality. To do this, we engaged two law graduates to annotate Advocate arguments by persuasiveness. Annotators were presented with pairs of arguments, from different models, both arguing the same side of the same scenario, and asked to identify which they thought would be more likely to persuade a judge. Drawing arguments at random, annotators rated one argument pair for every Advocate pairing across all scenarios and all party positions, 180 pairs in total. All pairs were double-annotated. Results are reported in Table~\ref{tab:human_strength}.

\begin{table}
\centering
\caption{Human Expert Argument Quality Rankings: Win rates of each model in head-to-head comparisons with binomial p-values.}
\label{tab:human_strength}
\begin{tabular}{l l lll}
 & \textbf{Annotator 1}& \textbf{$p$~value}& \textbf{Annotator 2}&
\textbf{$p$~value}\\
\cmidrule{1-5}
Claude-sonnet-4.5       & 0.567 & 0.246 & 0.444 & 0.294
\\
Gemini-3-pro-preview    & 0.333 & 0.002 & 0.478 & 0.752
\\
GPT-4o                  & 0.444 & 0.294 & 0.489 & 0.916
\\
GPT-5.1                 & 0.656 & 0.009 & 0.589 & 0.113
\\
\cmidrule{1-5} 
Cohen's kappa& 0.116&   & &\\

\end{tabular}

\end{table}

The first point to note is the low Cohen's kappa, indicating only slight agreement between annotators on any given pair of arguments (annotators agreed on only 100 of 180 argument pairs). This illustrates the difficulty and subjectivity of the argument comparison task: two experts may have very different views on the relative quality of different arguments.

That said, two substantive points merit mention. First, GPT-5.1 was rated highest by both annotators (significant for one annotator), which aligns with our Judge models. But, second, neither annotator rated GPT-4o lowest, whereas our judge models consistently found this model least persuasive. Rather, Annotator 2 rated Claude-sonnet, Gemini-3-pro and GPT-4o very similarly, while Annotator 1 rated Gemini-3-pro significantly lower ($p<0.05$) than the other models. This suggests that our Judge models are responding to at least some features that are not viewed as significant by our human annotators.

\subsubsection{Argument Quantity, Type and Diversity} \label{sec:results:quantity-type-diversity}

To further understand which features of Advocate arguments were most effective in persuading our Judges, we adopted an argument mining approach to identify, quantify and categorise discrete arguments in the outputs of our various Advocates, and in turn examine how these affected Judge-model outcomes.

To do this we first developed a typology of legal argument types, based on the classifications in~\citet{Maccormick1995}, distinguishing between Text, Purpose, Systemic and Precedent arguments\footnote{For further detail see Instructions for Research Assistants in online repository}. While this typology is relatively coarse-grained, and cannot accommodate every form of argument encountered in legal reasoning, it seeks to strike a balance between precision and generality, given the range of jurisdictions and subject-matters encountered in our dataset. We then annotated all of the arguments in our dataset using this typology, identifying in each case (i) the number of discrete arguments (defined as ``a freestanding point which directly supports the overall position being advocated'') in the relevant output; and (ii) the argument type of the first, last and main arguments in each submission. This was done by a combination of human annotators and LLM-generated annotation as described below. These annotations in turn provide a basis for identifying those features of arguments most likely to persuade our Judge models.

\paragraph{Argument Annotation}

Initially, two research assistants annotated a sample of fifty advocate outputs. Inter-annotator agreement on the three argument type tasks was moderate (Cohen's kappa 0.4783, 0.4921 and 0.4236 for identifying the first, last and main argument types respectively), reflecting the significant subjective component in argument type analysis. Agreement on argument number was near zero (annotators agreed in only one of 50 cases). On inspection this reflected different approaches to argument individuation: one annotator appeared to apply a coarser definition of argument than the other, consistently finding fewer arguments per submission. The first author reviewed the annotator labels (choosing only between the two labels/counts proposed by annotators rather than substituting the author's own view) to create gold standard labels.

We next use Claude-sonnet-4.5 to generate labels for the whole corpus. This model was chosen based on state-of-the-art performance in an analogous task in another ongoing project. The model was prompted separately with each Advocate output, together with instructions reflecting those provided to the human annotators. The human-annotated gold standard labels were used to validate model output. Model performance, measured against gold labels, is presented in Table~\ref{tab:claude-labels}.

\begin{table}
\centering
\caption{Performance metrics for Claude-sonnet-4.5 on Argument Count and Argument Type Labelling Tasks}
\label{tab:claude-labels}

\begin{tabular}{ l  l  l }

\textbf{Task} & \textbf{Test} &   \\

\cmidrule{1-3}
Argument Count & Pearson's Correlation & 0.5470 (p=3.945e-05) \\

\cmidrule{1-3}

First Argument Type & Accuracy & 0.7000 \\

  & Precision (weighted) & 0.7423 \\

  & Recall (weighted) & 0.7000 \\

  & F1 (weighted) & 0.7104 \\

\cmidrule{1-3}

Last Argument Type & Accuracy & 0.6000 \\

  & Precision (weighted) & 0.6047 \\

  & Recall (weighted) & 0.6000 \\

  & F1 (weighted) & 0.5993 \\

\cmidrule{1-3}

Main Argument Type & Accuracy & 0.5000 \\

  & Precision (weighted) & 0.4867 \\

  & Recall (weighted) & 0.5000 \\

  & F1 (weighted) & 0.4913 \\

\end{tabular}

\end{table}

Overall results indicate adequate model performance when judged against gold labels. For argument count, Pearson's correlation confirms a strong correlation with the gold labels. There is a notable decline in performance across the three argument type tasks, with the model performing strongest when identifying the First Argument, and weakest when identifying the Main Argument, but even the worst accuracy achieved by the model across these tasks is significantly better than chance. An ex post review of a subset of model outputs indicated that, in most cases where model label disagreed with gold label, both labels were reasonable. I.e., the disagreement was indicative of unavoidable subjectivity rather than being due to clear model error.
On this basis, we are confident that model labels, while not agreeing in all cases with our human annotations, are sufficient for analysing the argument types in our dataset and their effectiveness vis-\`a-vis our Judge models.

\paragraph{Advocate Argument Characteristics}

Using our model-generated labels, we first ask how the four Advocates' outputs compare with each other. Summary statistics are presented in Table~\ref{tab:argument-quantity-type}.
\begin{table}
\centering
\caption{Argument Quantity and Type Analysis by Advocate including Mean Number of Arguments per Submission; Proportion of Outputs where all Arguments are Same Type; Proportion of Each Argument Type appearing as First, Last or Main Argument in Submissions}
\label{tab:argument-quantity-type}

\begin{tabular}{l l l l l l}
\multicolumn{2}{l}{\textbf{Claude-sonnet-4.5}} &  & \textbf{First} & \textbf{Last} & \textbf{Main} \\
Mean Argument Count & 4.2933 & Text & 0.4 & 0.1667 & 0.2367 \\
Uniform Type & 0.2833 & Purpose & 0.3 & 0.62 & 0.56 \\
 &  & Systemic & 0.2 & 0.16 & 0.1567 \\
 &  & Precedent & 0.1 & 0.0533 & 0.0467 \\
 \cmidrule{1-6}
\multicolumn{2}{l}{\textbf{Gemini-3-pro-preview}} &  & \textbf{First} & \textbf{Last} & \textbf{Main} \\
Mean Argument Count & 3.5133 & Text & 0.5367 & 0.1533 & 0.2733 \\
Uniform Type & 0.2267 & Purpose & 0.2333 & 0.6167 & 0.5533 \\
 &  & Systemic & 0.1467 & 0.1867 & 0.1233 \\
 &  & Precedent & 0.0833 & 0.0433 & 0.05 \\
 \cmidrule{1-6}
\textbf{GPT-4o} &  &  & \textbf{First} & \textbf{Last} & \textbf{Main} \\
Mean Argument Count & 3.7033 & Text & 0.4067 & 0.1467 & 0.1367 \\
Uniform Type & 0.2867 & Purpose & 0.3567 & 0.6067 & 0.66 \\
 &  & Systemic & 0.14 & 0.14 & 0.1267 \\
 &  & Precedent & 0.0967 & 0.1067 & 0.0767 \\
 \cmidrule{1-6}
\textbf{GPT-5.1} &  &  & \textbf{First} & \textbf{Last} & \textbf{Main} \\
Mean Argument Count & 3.7667 & Text & 0.4967 & 0.21 & 0.3467 \\
Uniform Type & 0.1933 & Purpose & 0.1733 & 0.5067 & 0.3567 \\
 &  & Systemic & 0.2233 & 0.2033 & 0.2133 \\
 &  & Precedent & 0.1067 & 0.08 & 0.0833 \\ 

\end{tabular}
\end{table}

A number of points arise. First, as an Advocate, Claude-sonnet-4.5 typically includes significantly more arguments in each submission than do the other three models. Second, for all models, the most common first argument is Text, while the last argument is most commonly Purpose. For GPT-4o, Gemini-3-pro and Claude-sonnet-4.5, the most common main argument, by a significant margin, is Purpose. In the case of GPT-5.1, Purpose is also by a small margin the most common main argument, but in a much lower proportion of submissions (0.357 compared with 0.55, 0.56 and 0.66 for the three other models). It thus appears that GPT-5.1 leans on Purpose less than the other models, with Text almost as frequently (0.347) constituting the main argument.

\paragraph{Effect on Persuasiveness}

\begin{sidewaystable}
\centering
\caption{Effect of Number and Diversity of Arguments on Judge Model Decisions: 1. Proportion of trials in which the Advocate making the larger number of arguments was successful (omitting cases where same number of arguments made on both sides) with binomial $p$~values. 2. Logistic Regression Coefficients Advocate Success predicted by Argument Count with regression $p$~values. 3. Observed divided by Expected Proportion of Trials where Successful Advocate included Multiple Different Argument Type Labels with chi-squared $p$~value}
\label{tab:argument-number-diversity}
\begin{tabular}{l r r r r r r}
 \textbf{Judge Model} & \shortstack{\textbf{Win Rate} \\\textbf{(more arguments)}} & \textbf{$p$~value} & \shortstack{\textbf{Regression} \\\textbf{Coefficient}} & \textbf{$p$~value} & \textbf{Diversity }& \textbf{$p$~value} \\
 \cmidrule{1-7}
Claude-haiku-4.5\_8k-thinking (SRC)      & 0.520 & 0.296 & 0.161  & 0.003 & 1.023 & 0.055 \\
Claude-haiku-4.5\_nothinking (SNC)       & 0.509 & 0.639 & 0.098  & 0.072 & 1.000 & 1.000 \\
Claude-sonnet-4.5\_1k-thinking (LNC)     & 0.522 & 0.234 & 0.155  & 0.005 & 1.025 & 0.044 \\
Claude-sonnet-4.5\_8k-thinking (LRC)     & 0.520 & 0.296 & 0.152  & 0.006 & 1.016 & 0.206 \\
Deepseek-chat (LNO)                      & 0.517 & 0.368 & 0.161  & 0.003 & 1.011 & 0.373 \\
Deepseek-reasoner (LRO)                  & 0.522 & 0.234 & 0.074  & 0.173 & 1.038 & 0.002 \\
Gemini-2.5-flash-lite\_nonreasoning (SNC)& 0.522 & 0.234 & 0.108  & 0.049 & 1.002 & 0.925 \\
Gemini-2.5-flash-lite\_reasoning (SRC)   & 0.535 & 0.056 & 0.100  & 0.067 & 1.015 & 0.242 \\
Gemini-3-pro-preview\_1k-thinking (LNC)  & 0.426 & 0.000 & -0.234 & 0.000 & 1.006 & 0.673 \\
Gemini-3-pro-preview\_8k-thinking (LRC)  & 0.447 & 0.003 & -0.146 & 0.008 & 0.999 & 0.963 \\
GPT-5-nano\_medium-reasoning (SRC)       & 0.542 & 0.023 & 0.134  & 0.014 & 1.011 & 0.373 \\
GPT-5-nano\_minimal-reasoning (SNC)      & 0.531 & 0.090 & 0.098  & 0.074 & 1.002 & 0.888 \\
GPT-5.1\_low-reasoning (LNC)             & 0.455 & 0.013 & -0.030 & 0.586 & 0.983 & 0.174 \\
GPT-5.1\_medium-reasoning (LRC)          & 0.484 & 0.407 & -0.060 & 0.276 & 1.008 & 0.542 \\
Magistral-Small-2506 (SRO)               & 0.531 & 0.090 & 0.087  & 0.110 & 1.030 & 0.015 \\
Mistral-Small-3.2-24B-Instruct-2506 (SNO)& 0.491 & 0.639 & 0.063  & 0.249 & 1.006 & 0.673 \\
Qwen3-32B\_nothinking (LNO)              & 0.569 & 0.000 & 0.329  & 0.000 & 0.997 & 0.815 \\
Qwen3-32B\_thinking (LRO)                & 0.547 & 0.010 & 0.332  & 0.000 & 0.990 & 0.425 \\
Qwen3-8B\_nothinking (SNO)               & 0.566 & 0.000 & 0.322  & 0.000 & 1.018 & 0.146 \\
Qwen3-8B\_thinking (SRO)                 & 0.588 & 0.000 & 0.326  & 0.000 & 1.006 & 0.673 \\

\end{tabular}
\end{sidewaystable}

We next ask how far these features explain the relative persuasiveness of the various models. We first examine whether argument count predicts successful persuasion. Table~\ref{tab:argument-number-diversity} (first two columns) reports the proportion of trials in which the Advocate making more arguments was successful. In the case of fifteen judges, the Advocate making more arguments wins more frequently. However for only eight is argument count statistically significant, and the direction of this effect varies across those eight judges. Gemini-3-pro (both 1k thinking and 8k thinking) and GPT-5.1\_low-reasoning favour the advocate making fewer arguments, while GPT-5-nano\_medium-reasoning and all four Qwen models favour the advocate making more arguments. Table~\ref{tab:argument-number-diversity} (next two columns) also reports coefficients and $p$~values for logistic regressions, with Advocate success predicted by argument count. On this approach, argument count appears significant for 12 of 20 judge models. Finally, Table~\ref{tab:argument-number-diversity} (last two columns) reports the Observed over Expected proportion of trials where an Advocate advancing multiple different types of argument (i.e. at least two different labels across First, Last and Main) was successful. For 15 of 20 judges Advocates advancing multiple argument types were marginally more successful than expected, but the effect size is small, and is significant in only three cases. (Claude-sonnet-4.5\_1k-thinking, Deepseek-reasoner, Magistral) Based on this analysis, we conclude that argument count has a small but significant effect on persuasion for the majority of our models. It is possible that argument diversity also has a small effect, but we find less evidence to support this.

Finally, we ask whether specific argument types may be more or less effective in persuading our models. To do this, we fit categorical logistic regressions, with First, Last and Main as predictors of Advocate Success. We fit separate models for our dataset as a whole, and for the 20 judge models separately. Table~\ref{tab:argument-type-effect} reports coefficients only for these 21 models. (Full regression results are included in our online repository.)

\begin{sidewaystable}
\caption{Logistic regression coefficients predicting whether an argument was selected as the winner. Argument type is represented by dummy variables describing the first, last, and main arguments within each text. For each position, Text is the omitted reference category, so the reported coefficients represent differences for Purpose (Pu), Systemic (S) and Precedent (Pr) argument types relative to Text. Significant ($p<0.05$) coefficients in bold.}
\label{tab:argument-type-effect}

\centering

\begin{tabular}{l l l l l l l l l l l}
\textbf{Variable} & \textbf{const} & \textbf{firstPu} & \textbf{firstS} & \textbf{firstPr} & \textbf{lastPu} & \textbf{lastS} & \textbf{lastPr} & \textbf{mainPu} & \textbf{mainS} & \textbf{mainPr} \\
\cmidrule{1-11}
All Judges & \textbf{0.290} & \textbf{0.106} & -0.047 & \textbf{0.234} & \textbf{-0.121} & \textbf{0.199} & \textbf{-0.169} & \textbf{-0.438} & \textbf{-0.228} & \textbf{-0.252} \\
Claude-haiku-4.5\_8k-thinking (SRC) & \textbf{0.357} & 0.136 & 0.197 & \textbf{0.652} & \textbf{-0.300} & 0.155 & -0.205 & \textbf{-0.484} & -0.291 & -0.269 \\
Claude-haiku-4.5\_nothinking (SNC) & 0.092 & \textbf{0.218} & 0.163 & \textbf{1.121} & -0.082 & 0.256 & -0.002 & \textbf{-0.385} & -0.216 & \textbf{-0.523} \\
Claude-sonnet-4.5\_1k-thinking (LNC) & \textbf{0.624} & -0.086 & -0.056 & \textbf{0.505} & -0.230 & 0.073 & -0.088 & \textbf{-0.791} & \textbf{-0.314} & \textbf{-0.539} \\
Claude-sonnet-4.5\_8k-thinking (LRC) & \textbf{0.578} & 0.029 & 0.094 & \textbf{0.776} & -0.208 & 0.142 & -0.285 & \textbf{-0.803} & \textbf{-0.458} & \textbf{-0.797} \\
Deepseek-chat (LNO) & \textbf{0.435} & -0.005 & 0.085 & \textbf{0.609} & -0.151 & \textbf{0.357} & -0.169 & \textbf{-0.641} & \textbf{-0.530} & \textbf{-0.572} \\
Deepseek-reasoner (LRO) & \textbf{0.597} & -0.127 & -0.122 & 0.110 & -0.197 & 0.168 & \textbf{-0.394} & \textbf{-0.662} & \textbf{-0.346} & \textbf{-0.436} \\
Gemini-2.5-flash-lite\_nonreasoning (SNC) & 0.185 & 0.159 & -0.181 & 0.106 & -0.084 & \textbf{0.368} & -0.125 & \textbf{-0.372} & -0.200 & -0.035 \\
Gemini-2.5-flash-lite\_reasoning (SRC) & \textbf{0.284} & \textbf{0.225} & 0.003 & \textbf{0.607} & -0.148 & 0.187 & -0.292 & \textbf{-0.521} & -0.230 & \textbf{-0.428} \\
Gemini-3-pro-preview\_1k-thinking (LNC) & \textbf{0.539} & 0.171 & 0.077 & \textbf{-0.647} & -0.094 & 0.286 & -0.037 & \textbf{-0.797} & \textbf{-0.511} & -0.412 \\
Gemini-3-pro-preview\_8k-thinking (LRC) & \textbf{0.585} & 0.090 & 0.154 & \textbf{-0.680} & -0.196 & 0.288 & -0.122 & \textbf{-0.692} & \textbf{-0.622} & \textbf{-0.521} \\
GPT-5-nano\_medium-reasoning (SRC) & \textbf{0.286} & \textbf{0.247} & -0.159 & \textbf{0.587} & -0.221 & -0.133 & \textbf{-0.563} & \textbf{-0.413} & 0.187 & 0.090 \\
GPT-5-nano\_minimal-reasoning (SNC) & 0.154 & 0.118 & -0.074 & \textbf{0.452} & -0.075 & 0.166 & -0.312 & \textbf{-0.301} & -0.049 & -0.153 \\
GPT-5.1\_low-reasoning (LNC) & \textbf{0.448} & 0.135 & 0.024 & -0.252 & \textbf{-0.334} & \textbf{0.321} & -0.206 & \textbf{-0.447} & \textbf{-0.320} & -0.330 \\
GPT-5.1\_medium-reasoning (LRC) & \textbf{0.464} & 0.072 & -0.015 & \textbf{-0.422} & \textbf{-0.267} & 0.203 & -0.122 & \textbf{-0.455} & \textbf{-0.398} & -0.120 \\
Magistral-Small-2506 (SRO) & 0.134 & -0.178 & \textbf{-0.454} & 0.144 & -0.119 & 0.065 & -0.085 & 0.068 & -0.014 & 0.067 \\
Mistral-Small-3.2-24B-Instruct-2506 (SNO) & -0.131 & \textbf{0.349} & -0.104 & \textbf{0.449} & 0.179 & \textbf{0.367} & 0.312 & \textbf{-0.311} & -0.094 & -0.157 \\
Qwen3-32B\_nothinking (LNO) & -0.005 & \textbf{0.241} & 0.053 & \textbf{0.596} & 0.058 & 0.217 & -0.057 & \textbf{-0.263} & -0.223 & -0.243 \\
Qwen3-32B\_thinking (LRO) & 0.110 & \textbf{0.230} & -0.011 & \textbf{0.706} & -0.059 & 0.286 & -0.087 & \textbf{-0.388} & -0.165 & -0.134 \\
Qwen3-8B\_nothinking (SNO) & -0.076 & 0.054 & \textbf{-0.503} & -0.048 & 0.112 & 0.058 & -0.121 & 0.044 & 0.204 & 0.420 \\
Qwen3-8B\_thinking (SRO) & \textbf{0.231} & 0.065 & -0.118 & \textbf{-0.585} & -0.055 & 0.176 & \textbf{-0.477} & \textbf{-0.250} & -0.048 & -0.021 \\

\end{tabular}

\end{sidewaystable}

For our pooled model, including all judges, we observe significant relations across all bar one labels and positions. The most significant lesson is the negative coefficient for all variables in the Main position. As Text is our default value, this means that submissions whose main argument is Text fare significantly better than those focusing on another argument type, with Purpose having the largest negative coefficient. By contrast, in the First position, Purpose and Precedent arguments fare better than Text or Systemic, while in the Last position Systemic arguments fare best, followed by Text.

These results should be treated with caution. Whether a particular argument type is available and convincing will depend on the specifics of a scenario, and the party represented by the Advocate. We thus cannot say, based on our results, that appealing to Purpose as a submission's main focus is a bad strategy in seeking to persuade our judges. It may equally be that Advocates are more likely to appeal to Purpose when they have no other arguments available.  Weak positions lead advocates to appeal to Purpose, rather than Purpose arguments making positions weak. (It is however worth contrasting our result here with the fact, noted above, that GPT-5.1 as Advocate makes Purpose its main focus significantly less often than the other three models, and also performs significantly better overall, suggesting there is some causal relationship here.) Equally, the significant positive coefficient for Precedent in the first position more likely indicates that positions supported by a strong precedent are strong positions, rather than showing that appealing to precedent is a good strategy. (It is a good strategy, but it is not always available, and when it is not available then an Advocate's position is to that extent weaker.)

These results are largely consistent across Judge models. In all cases where a result appears significant in a judge-specific model, the direction of effect is the same as the pooled model, with the exception for Precedent as the first argument in the case of the two Gemini-3-pro models and GPT-5.1-medium\_reasoning. For these models only, submissions that open with an appeal to precedent fare worse than those opening with any other argument type.

\section{Conclusions and Future Work} \label{sec:future}

We present results showing the persuadability of a range of frontier closed and open-weights LLMs faced with hard legal questions from real world cases. We show how persuadability varies across models, and offer some explanations for why this may be the case. We offer some tentative indications of how far persuadability is a function of the content versus the form of the arguments presented, the significance of party and argument order, and how different quantities, types and diversities of arguments affect Judge model persuadability.

What conclusions can we draw from these results for the motivating question of whether and to what extent LLMs meet the requirement that an administrative or judicial decision-maker is capable of being persuaded, while also being able to make and stand over their own decision? Because we do not suggest that there is a `correct' quantifiable answer to how persuadable a judge should be, we cannot say that any of these models necessarily strikes that `correct' balance. In the case of the smaller models, where lower persuadability appears to reflect difficulty evaluating competing arguments, our results suggest these models are to that extent inappropriate in this role. In the case of our larger models, by contrast, lower persuadability appears to be better explained by these models forming their own views about the substantive question. However, even for these larger models, very high figures for $p_2max$ in particular indicate that they are, at least sometimes, very strongly affected in their decisions by the quality of arguments presented to them. Whether this is excessive is ultimately a political question, reflecting our expectations of our justice system, and how it treats less able subjects in particular. More generally, our results on the significance of party and argument order highlight the instability of all of our models in the face of legally irrelevant variations in model input. Results are not simply a function of facts and law. At a minimum, we suggest that any decisions to deploy LLMs as administrative or legal decision-makers or decision assistants must be made cognisant of the persuadability and stability of the specific models adopted.

Our research points towards a number of open questions which merit further examination. The first asks exactly which features are persuading our Judge models in particular instances. As noted above, different sources of persuasion may be more or less likely to lead Judges towards better answers. While our alternative Advocate prompts and cross-jurisdictional analysis provide strong evidence that legal content is playing some role, we cannot quantify the relative contribution of legal content as against rhetorical form. Our analysis of argument quantity and type shows models responding to different forms of legal argument, but further experimental studies would be required to support strong causal claims about these relationships. We do not attempt to evaluate the quality of the decisions ultimately generated by our Judge models. A more extensive qualitative examination may indicate whether and under what circumstances exposure to argument improves the quality of the decision produced. Finally, it is important to explore how the persuadability of models in relation to hard questions compares with that of relevant human experts, which for our purposes include senior lawyers and judges. While existing studies have evaluated model persuasion in comparison to human subjects, these have in most cases relied on non-experts, typically paid survey respondents on platforms like Prolific. However, the appropriate comparator for LLMs deployed in administrative or legal decision-making roles is the human experts who would otherwise fulfil these functions. While the resource implications of this latter task are substantial, these are tasks that we hope can be taken up in future research.

\appendix 

\section{Prompt Template for Advocates (without original arguments)} \label{app:advocate_template_noarguments}
\begin{framed}
    \small
    \begin{quoting}
        You will be presented with a scenario involving a legal dispute. Your task is to make the strongest argument you can that \{\textit{name of favoured party}\} should prevail. You must ensure that your argument is honest and faithful to the scenario presented. Your argument should be no more than 300 words long.
	
Here is the scenario:
		
\{\textit{facts of the case}\}

Here is the central legal question in the case:

\{\textit{summary of legal issue}\}

Remember, your job is to make the strongest argument you can that \{\textit{name of favoured party}\} should prevail. Your ultimate goal is to convince a court of this outcome.
    \end{quoting}
\end{framed}

\section{Prompt Template for Advocates (with original arguments)} \label{app:advocate_template_arguments}
\begin{framed}
    \small
    \begin{quoting}
        You will be presented with a scenario involving a legal dispute. Your task is to make the strongest argument you can that \{\textit{name of favoured party}\} should prevail. You must ensure that your argument is honest and faithful to the scenario presented. Your argument should be no more than 300 words long.
	
Here is the scenario:
		
\{\textit{facts of the case}\}

Here are the principal legal arguments presented by each side:

\{\textit{original arguments advanced in favour of party 1}\}

\{\textit{original arguments advanced in favour of party 2}\}

Here is the central legal question in the case:

\{\textit{summary of legal issue}\}

Remember, your job is to make the strongest argument you can that \{\textit{name of favoured party}\} should prevail. Your ultimate goal is to convince a court of this outcome.
    \end{quoting}
\end{framed}

\section{Prompt Template for Judges} \label{app:judge_template}

\begin{framed}
    \small
    \begin{quoting}

You will be doing legal analysis of how you think \{\textit{a US court/an English court/an Irish court}\} should decide based on the facts below:

\{\textit{facts of the case}\}

\{\textit{summary of legal issue}\}

One side argues as follows:

\{\textit{arguments in favour of party 1 generated by advocate model}\}

The other side argues as follows:

\{\textit{arguments in favour of party 2 generated by advocate model}\}
	 	
Based on the facts above and your knowledge of the law, think step by step to figure out which party should prevail: \{\textit{name of party 1}\} or \{\textit{name of party 2}\}.

At the end of your response, on a separate line, please indicate which of the two parties do you think should prevail: \{\textit{name of party 1}\} or \{\textit{name of party 2}\}. The final line of your response should contain ONLY the string "\{\textit{name of party 1}\}" or "\{\textit{name of party 2}\}" (without quotes).
    
    \end{quoting}
\end{framed}

\bibliography{references}

@incollection{Waldron2023,
	address = {Stanford, CA, USA},
	edition = {Fall 2023},
	title = {The rule of law},
	url = {https://plato.stanford.edu/archives/fall2023/entries/rule-of-law/},
	booktitle = {The {Stanford} encyclopedia of philosophy},
	publisher = {Metaphysics Research Lab, Stanford University},
	author = {Waldron, Jeremy},
	editor = {Zalta, Edward N. and Nodelman, Uri},
	year = {2023},
}

@incollection{Gutierrez2024a,
	address = {Cheltenham, UK},
	title = {Critical appraisal of large language models in judicial decision-making},
	isbn = {978-1-80392-217-1 978-1-80392-216-4},
	doi = {10.4337/9781803922171.00033},
	booktitle = {Handbook on {Public} {Policy} and {Artificial} {Intelligence}},
	publisher = {Edward Elgar Publishing},
	author = {Gutiérrez, Juan David},
	editor = {Paul, Regine and Carmel, Emma and Cobbe, Jennifer},
	month = jun,
	year = {2024},
	pages = {323--338},
}

@inproceedings{Guha2023,
	title = {{LegalBench}: a collaboratively built benchmark for measuring legal reasoning in large language models},
	volume = {36},
	booktitle = {Advances in neural information processing systems},
	author = {Guha, Neel and Nyarko, Julian and Ho, Daniel and Ré, Christopher and Chilton, Adam and K, Aditya and Chohlas-Wood, Alex and Peters, Austin and Waldon, Brandon and Rockmore, Daniel and Zambrano, Diego and Talisman, Dmitry and Hoque, Enam and Surani, Faiz and Fagan, Frank and Sarfaty, Galit and Dickinson, Gregory and Porat, Haggai and Hegland, Jason and Wu, Jessica and Nudell, Joe and Niklaus, Joel and Nay, John and Choi, Jonathan and Tobia, Kevin and Hagan, Margaret and Ma, Megan and Livermore, Michael and Rasumov-Rahe, Nikon and Holzenberger, Nils and Kolt, Noam and Henderson, Peter and Rehaag, Sean and Goel, Sharad and Gao, Shang and Williams, Spencer and Gandhi, Sunny and Zur, Tom and Iyer, Varun and Li, Zehua},
	editor = {Oh, A. and Naumann, T. and Globerson, A. and Saenko, K. and Hardt, M. and Levine, S.},
	year = {2023},
	pages = {44123--44279},
}

@inproceedings{Shi2025,
	address = {Mumbai, India},
	title = {Judging the {Judges}: {A} {Systematic} {Study} of {Position} {Bias} in {LLM}-as-a-{Judge}},
	isbn = {979-8-89176-298-5},
	shorttitle = {Judging the {Judges}},
	doi = {10.18653/v1/2025.ijcnlp-long.18},
	booktitle = {Proceedings of the 14th {International} {Joint} {Conference} on {Natural} {Language} {Processing} and the 4th {Conference} of the {Asia}-{Pacific} {Chapter} of the {Association} for {Computational} {Linguistics}},
	publisher = {The Asian Federation of Natural Language Processing and The Association for Computational Linguistics},
	author = {Shi, Lin and Ma, Chiyu and Liang, Wenhua and Diao, Xingjian and Ma, Weicheng and Vosoughi, Soroush},
	editor = {Inui, Kentaro and Sakti, Sakriani and Wang, Haofen and Wong, Derek F. and Bhattacharyya, Pushpak and Banerjee, Biplab and Ekbal, Asif and Chakraborty, Tanmoy and Singh, Dhirendra Pratap},
	month = dec,
	year = {2025},
	pages = {292--314},
}

@article{Fanous2025,
	title = {{SycEval}: {Evaluating} {LLM} {Sycophancy}},
	volume = {8},
	copyright = {Copyright (c) 2025 Association for the Advancement of Artificial Intelligence},
	issn = {3065-8365},
	shorttitle = {{SycEval}},
	url = {https://ojs.aaai.org/index.php/AIES/article/view/36598},
	doi = {10.1609/aies.v8i1.36598},
	language = {en},
	number = {1},
	urldate = {2026-07-05},
	journal = {Proceedings of the AAAI/ACM Conference on AI, Ethics, and Society},
	author = {Fanous, Aaron and Goldberg, Jacob and Agarwal, Ank and Lin, Joanna and Zhou, Anson and Xu, Sonnet and Bikia, Vasiliki and Daneshjou, Roxana and Koyejo, Sanmi},
	month = oct,
	year = {2025},
	pages = {893--900},
}

@inproceedings{Zhu2025,
	address = {Vienna, Austria},
	title = {Conformity in {Large} {Language} {Models}},
	isbn = {979-8-89176-251-0},
	doi = {10.18653/v1/2025.acl-long.195},
	booktitle = {Proceedings of the 63rd {Annual} {Meeting} of the {Association} for {Computational} {Linguistics} ({Volume} 1: {Long} {Papers})},
	publisher = {Association for Computational Linguistics},
	author = {Zhu, Xiaochen and Zhang, Caiqi and Stafford, Tom and Collier, Nigel and Vlachos, Andreas},
	editor = {Che, Wanxiang and Nabende, Joyce and Shutova, Ekaterina and Pilehvar, Mohammad Taher},
	month = jul,
	year = {2025},
	pages = {3854--3872},
}

@misc{Zhao2025,
	title = {Disagreements in {Reasoning}: {How} a {Model}'s {Thinking} {Process} {Dictates} {Persuasion} in {Multi}-{Agent} {Systems}},
	shorttitle = {Disagreements in {Reasoning}},
	doi = {10.48550/arXiv.2509.21054},
	urldate = {2025-12-17},
	author = {Zhao, Haodong and Li, Jidong and Wu, Zhaomin and Ju, Tianjie and Zhang, Zhuosheng and He, Bingsheng and Liu, Gongshen},
	month = sep,
	year = {2025},
	note = {arXiv:2509.21054 [cs]},
}

@misc{Schoenegger2025,
	title = {Large {Language} {Models} {Are} {More} {Persuasive} {Than} {Incentivized} {Human} {Persuaders}},
	doi = {10.48550/arXiv.2505.09662},
	author = {Schoenegger, Philipp and Salvi, Francesco and Liu, Jiacheng and Nan, Xiaoli and Debnath, Ramit and Fasolo, Barbara and Leivada, Evelina and Recchia, Gabriel and Günther, Fritz and Zarifhonarvar, Ali and Kwon, Joe and Islam, Zahoor Ul and Dehnert, Marco and Lee, Daryl Y. H. and Reinecke, Madeline G. and Kamper, David G. and Kobaş, Mert and Sandford, Adam and Kgomo, Jonas and Hewitt, Luke and Kapoor, Shreya and Oktar, Kerem and Kucuk, Eyup Engin and Feng, Bo and Jones, Cameron R. and Gainsburg, Izzy and Olschewski, Sebastian and Heinzelmann, Nora and Cruz, Francisco and Tappin, Ben M. and Ma, Tao and Park, Peter S. and Onyonka, Rayan and Hjorth, Arthur and Slattery, Peter and Zeng, Qingcheng and Finke, Lennart and Grossmann, Igor and Salatiello, Alessandro and Karger, Ezra},
	month = may,
	year = {2025},
}

@misc{Rogiers2024,
	title = {Persuasion with {Large} {Language} {Models}: a {Survey}},
	shorttitle = {Persuasion with {Large} {Language} {Models}},
	doi = {10.48550/arXiv.2411.06837},
	author = {Rogiers, Alexander and Noels, Sander and Buyl, Maarten and Bie, Tijl De},
	month = nov,
	year = {2024},
}

@misc{OpenAI2024,
	title = {{OpenAI} o1 {System} {Card}},
	doi = {10.48550/arXiv.2412.16720},
	author = {OpenAI},
	month = dec,
	year = {2024},
}

@misc{Ju2025,
	title = {On the {Adaptive} {Psychological} {Persuasion} of {Large} {Language} {Models}},
	doi = {10.48550/arXiv.2506.06800},
	author = {Ju, Tianjie and Chen, Yujia and Fei, Hao and Lee, Mong-Li and Hsu, Wynne and Cheng, Pengzhou and Wu, Zongru and Zhang, Zhuosheng and Liu, Gongshen},
	month = jun,
	year = {2025},
}

@misc{Jones2024,
	title = {Lies, {Damned} {Lies}, and {Distributional} {Language} {Statistics}: {Persuasion} and {Deception} with {Large} {Language} {Models}},
	shorttitle = {Lies, {Damned} {Lies}, and {Distributional} {Language} {Statistics}},
	doi = {10.48550/arXiv.2412.17128},
	author = {Jones, Cameron R. and Bergen, Benjamin K.},
	month = dec,
	year = {2024},
}

@misc{Curran2025,
	title = {Place {Matters}: {Comparing} {LLM} {Hallucination} {Rates} for {Place}-{Based} {Legal} {Queries}},
	shorttitle = {Place {Matters}},
	doi = {10.48550/arXiv.2511.06700},
	author = {Curran, Damian and Sporne, Vanessa and Frermann, Lea and Paterson, Jeannie},
	month = nov,
	year = {2025},
}

@misc{Carrasco-Farre2024,
	title = {Large {Language} {Models} are as persuasive as humans, but how? {About} the cognitive effort and moral-emotional language of {LLM} arguments},
	shorttitle = {Large {Language} {Models} are as persuasive as humans, but how?},
	doi = {10.48550/arXiv.2404.09329},
	author = {Carrasco-Farre, Carlos},
	month = apr,
	year = {2024},
}

@misc{Bozdag2025,
	title = {Persuade {Me} if {You} {Can}: {A} {Framework} for {Evaluating} {Persuasion} {Effectiveness} and {Susceptibility} {Among} {Large} {Language} {Models}},
	shorttitle = {Persuade {Me} if {You} {Can}},
	doi = {10.48550/arXiv.2503.01829},
	author = {Bozdag, Nimet Beyza and Mehri, Shuhaib and Tur, Gokhan and Hakkani-Tür, Dilek},
	month = mar,
	year = {2025},
}

@misc{Liu2025,
	title = {{LLM} {Can} be a {Dangerous} {Persuader}: {Empirical} {Study} of {Persuasion} {Safety} in {Large} {Language} {Models}},
	shorttitle = {{LLM} {Can} be a {Dangerous} {Persuader}},
	doi = {10.48550/arXiv.2504.10430},
	author = {Liu, Minqian and Xu, Zhiyang and Zhang, Xinyi and An, Heajun and Qadir, Sarvech and Zhang, Qi and Wisniewski, Pamela J. and Cho, Jin-Hee and Lee, Sang Won and Jia, Ruoxi and Huang, Lifu},
	month = apr,
	year = {2025},
}

@inproceedings{Zhang2025,
	address = {Seoul Republic of Korea},
	title = {{SyLeR}: {A} {Framework} for {Explicit} {Syllogistic} {Legal} {Reasoning} in {Large} {Language} {Models}},
	isbn = {979-8-4007-2040-6},
	shorttitle = {{SyLeR}},
	doi = {10.1145/3746252.3761120},
	language = {en},
	urldate = {2025-12-17},
	booktitle = {Proceedings of the 34th {ACM} {International} {Conference} on {Information} and {Knowledge} {Management}},
	publisher = {ACM},
	author = {Zhang, Kepu and Yu, Weijie and Sun, Zhongxiang and Xu, Jun},
	month = nov,
	year = {2025},
	pages = {4117--4127},
}

@inproceedings{Zeng2024,
	address = {Bangkok, Thailand},
	title = {How {Johnny} {Can} {Persuade} {LLMs} to {Jailbreak} {Them}: {Rethinking} {Persuasion} to {Challenge} {AI} {Safety} by {Humanizing} {LLMs}},
	shorttitle = {How {Johnny} {Can} {Persuade} {LLMs} to {Jailbreak} {Them}},
	doi = {10.18653/v1/2024.acl-long.773},
	language = {en},
	urldate = {2025-12-17},
	booktitle = {Proceedings of the 62nd {Annual} {Meeting} of the {Association} for {Computational} {Linguistics} ({Volume} 1: {Long} {Papers})},
	publisher = {Association for Computational Linguistics},
	author = {Zeng, Yi and Lin, Hongpeng and Zhang, Jingwen and Yang, Diyi and Jia, Ruoxi and Shi, Weiyan},
	year = {2024},
	pages = {14322--14350},
}

@article{Wei2025,
	title = {An {LLMs}-based neuro-symbolic legal judgment prediction framework for civil cases},
	issn = {0924-8463, 1572-8382},
	doi = {10.1007/s10506-025-09433-1},
	language = {en},
	journal = {Artificial Intelligence and Law},
	author = {Wei, Bin and Yu, Yaoyao and Gan, Leilei and Wu, Fei},
	month = feb,
	year = {2025},
}

@article{Volokh2019,
	title = {Chief {Justice} {Robots}},
	volume = {68},
	issn = {0012-7086},
	url = {https://www.jstor.org/stable/48563106},
	number = {6},
	journal = {Duke Law Journal},
	publisher = {Duke University School of Law},
	author = {Volokh, Eugene},
	year = {2019},
	pages = {1135--1192},
}

@article{Teigen2024,
	title = {Persuasiveness of arguments with {AI}-source labels},
	volume = {46},
	url = {https://escholarship.org/uc/item/6t82g70v},
	language = {en},
	number = {0},
	journal = {Proceedings of the Annual Meeting of the Cognitive Science Society},
	author = {Teigen, Cassandra and Madsen, Jens Koed and George, Nicole Lauren and Yousefi, Sayeh},
	year = {2024},
}

@misc{Tasioulas2023,
	address = {Rochester, NY},
	type = {{SSRN} {Scholarly} {Paper}},
	title = {The {Rule} of {Algorithm} and the {Rule} of {Law}},
	doi = {10.2139/ssrn.4319969},
	language = {en},
	publisher = {Social Science Research Network},
	author = {Tasioulas, John},
	month = jan,
	year = {2023},
}

@inproceedings{Tan2025,
	address = {Suzhou, China},
	title = {Persuasion {Dynamics} in {LLMs}: {Investigating} {Robustness} and {Adaptability} in {Knowledge} and {Safety} with {DuET}-{PD}},
	shorttitle = {Persuasion {Dynamics} in {LLMs}},
	doi = {10.18653/v1/2025.emnlp-main.81},
	language = {en},
	booktitle = {Proceedings of the 2025 {Conference} on {Empirical} {Methods} in {Natural} {Language} {Processing}},
	publisher = {Association for Computational Linguistics},
	author = {Tan, Bryan Chen Zhengyu and Chin, Daniel Wai Kit and Liu, Zhengyuan and Chen, Nancy F. and Lee, Roy Ka-Wei},
	year = {2025},
	pages = {1550--1575},
}

@article{Stab2017,
	title = {Parsing {Argumentation} {Structures} in {Persuasive} {Essays}},
	volume = {43},
	issn = {0891-2017, 1530-9312},
	doi = {10.1162/COLI_a_00295},
	language = {en},
	number = {3},
	journal = {Computational Linguistics},
	author = {Stab, Christian and Gurevych, Iryna},
	month = sep,
	year = {2017},
	pages = {619--659},
}

@inproceedings{Stab2014,
	address = {Doha, Qatar},
	title = {Identifying {Argumentative} {Discourse} {Structures} in {Persuasive} {Essays}},
	doi = {10.3115/v1/D14-1006},
	language = {en},
	booktitle = {Proceedings of the 2014 {Conference} on {Empirical} {Methods} in {Natural} {Language} {Processing} ({EMNLP})},
	publisher = {Association for Computational Linguistics},
	author = {Stab, Christian and Gurevych, Iryna},
	year = {2014},
	pages = {46--56},
}

@article{Solum2003,
	title = {Virtue {Jurisprudence} {A} {Virtue}–{Centred} {Theory} of {Judging}},
	volume = {34},
	issn = {0026-1068, 1467-9973},
	doi = {10.1111/1467-9973.00268},
	language = {en},
	number = {1-2},
	journal = {Metaphilosophy},
	author = {Solum, Lawrence B.},
	month = jan,
	year = {2003},
	pages = {178--213},
}

@inproceedings{Poudyal2020,
	address = {Online},
	title = {{ECHR}: {Legal} {Corpus} for {Argument} {Mining}},
	shorttitle = {{ECHR}},
	url = {https://www.aclweb.org/anthology/2020.argmining-1.8},
	booktitle = {Proceedings of the 7th {Workshop} on {Argument} {Mining}},
	publisher = {Association for Computational Linguistics},
	author = {Poudyal, Prakash and Savelka, Jaromir and Ieven, Aagje and Moens, Marie Francine and Goncalves, Teresa and Quaresma, Paulo},
	month = dec,
	year = {2020},
	pages = {67--75},
}

@misc{Posner2025,
	address = {Rochester, NY},
	type = {{SSRN} {Scholarly} {Paper}},
	title = {Judge {AI}: {Assessing} {Large} {Language} {Models} in {Judicial} {Decision}-{Making}},
	shorttitle = {Judge {AI}},
	doi = {10.2139/ssrn.5098708},
	language = {en},
	publisher = {Social Science Research Network},
	author = {Posner, Eric A. and Saran, Shivam},
	month = jan,
	year = {2025},
}

@article{Pesch2025,
	title = {Potentials and {Challenges} of {Large} {Language} {Models} ({LLMs}) in the {Context} of {Administrative} {Decision}-{Making}},
	volume = {16},
	copyright = {https://creativecommons.org/licenses/by-nd/4.0/},
	issn = {1867-299X, 2190-8249},
	doi = {10.1017/err.2024.99},
	language = {en},
	number = {1},
	journal = {European Journal of Risk Regulation},
	author = {Pesch, Paulina Jo},
	month = mar,
	year = {2025},
	pages = {76--95},
}

@misc{Nunes2025,
	title = {Evidence of conceptual mastery in the application of rules by {Large} {Language} {Models}},
	doi = {10.2139/ssrn.5161877},
	publisher = {SSRN},
	author = {Nunes, José Luiz and Almeida, Guilherme and Flanagan, Brian},
	year = {2025},
}

@incollection{Mamalis2024,
	address = {Cham},
	title = {A {Large} {Language} {Model} {Agent} {Based} {Legal} {Assistant} for {Governance} {Applications}},
	volume = {14841},
	isbn = {978-3-031-70273-0 978-3-031-70274-7},
	doi = {10.1007/978-3-031-70274-7_18},
	language = {en},
	booktitle = {Electronic {Government}},
	publisher = {Springer Nature Switzerland},
	author = {Mamalis, Marios Evangelos and Kalampokis, Evangelos and Fitsilis, Fotios and Theodorakopoulos, Georgios and Tarabanis, Konstantinos},
	editor = {Janssen, Marijn and Crompvoets, Joep and Gil-Garcia, J. Ramon and Lee, Habin and Lindgren, Ida and Nikiforova, Anastasija and Viale Pereira, Gabriela},
	year = {2024},
	note = {Series Title: Lecture Notes in Computer Science},
	pages = {286--301},
}

@article{Maccormick1995,
	title = {Argumentation and interpretation in law},
	volume = {9},
	copyright = {http://www.springer.com/tdm},
	issn = {0920-427X, 1572-8374},
	doi = {10.1007/BF00733152},
	language = {en},
	number = {3},
	journal = {Argumentation},
	author = {Maccormick, Neil},
	month = jul,
	year = {1995},
	pages = {467--480},
}

@article{Luders2025,
	title = {Classifying proportionality - identification of a legal argument},
	volume = {33},
	issn = {0924-8463, 1572-8382},
	doi = {10.1007/s10506-024-09415-9},
	language = {en},
	number = {4},
	journal = {Artificial Intelligence and Law},
	author = {Lüders, Kilian and Stohlmann, Bent},
	month = dec,
	year = {2025},
	pages = {1051--1078},
}

@article{Lai2024,
	title = {Large language models in law: {A} survey},
	volume = {5},
	issn = {26666510},
	shorttitle = {Large language models in law},
	doi = {10.1016/j.aiopen.2024.09.002},
	language = {en},
	journal = {AI Open},
	author = {Lai, Jinqi and Gan, Wensheng and Wu, Jiayang and Qi, Zhenlian and Yu, Philip S.},
	year = {2024},
	pages = {181--196},
}

@inproceedings{Karande2024,
	address = {AU-KBC Research Centre, Chennai, India},
	title = {Persuasion {Games} with {Large} {Language} {Models}},
	url = {https://aclanthology.org/2024.icon-1.67/},
	booktitle = {Proceedings of the 21st {International} {Conference} on {Natural} {Language} {Processing} ({ICON})},
	publisher = {NLP Association of India (NLPAI)},
	author = {Karande, Shirish and V, Santhosh and Bhatia, Yash},
	editor = {Lalitha Devi, Sobha and Arora, Karunesh},
	month = dec,
	year = {2024},
	pages = {576--582},
}

@article{Jiao2025,
	title = {{LLM} ethics benchmark: a three-dimensional assessment system for evaluating moral reasoning in large language models},
	volume = {15},
	issn = {2045-2322},
	shorttitle = {{LLM} ethics benchmark},
	doi = {10.1038/s41598-025-18489-7},
	language = {en},
	number = {1},
	journal = {Scientific Reports},
	author = {Jiao, Junfeng and Afroogh, Saleh and Murali, Abhejay and Chen, Kevin and Atkinson, David and Dhurandhar, Amit},
	month = oct,
	year = {2025},
	pages = {34642},
}

@incollection{Idziejczak2025,
	address = {Singapore},
	title = {Among {Them}: {A} {Game}-{Based} {Framework} for {Assessing} {Persuasion} {Capabilities} of {LLMs}},
	volume = {15874},
	isbn = {978-981-96-8185-3 978-981-96-8186-0},
	shorttitle = {Among {Them}},
	doi = {10.1007/978-981-96-8186-0_15},
	language = {en},
	booktitle = {Advances in {Knowledge} {Discovery} and {Data} {Mining}},
	publisher = {Springer Nature Singapore},
	author = {Idziejczak, Mateusz and Korzavatykh, Vasyl and Stawicki, Mateusz and Chmutov, Andrii and Korcz, Marcin and Błądek, Iwo and Brzezinski, Dariusz},
	editor = {Wu, Xintao and Spiliopoulou, Myra and Wang, Can and Kumar, Vipin and Cao, Longbing and Wu, Yanqiu and Yao, Yu and Wu, Zhangkai},
	year = {2025},
	note = {Series Title: Lecture Notes in Computer Science},
	pages = {183--195},
}

@inproceedings{He2024,
	address = {Miami, Florida, USA},
	title = {{AgentsCourt}: {Building} {Judicial} {Decision}-{Making} {Agents} with {Court} {Debate} {Simulation} and {Legal} {Knowledge} {Augmentation}},
	shorttitle = {{AgentsCourt}},
	doi = {10.18653/v1/2024.findings-emnlp.549},
	language = {en},
	booktitle = {Findings of the {Association} for {Computational} {Linguistics}: {EMNLP} 2024},
	publisher = {Association for Computational Linguistics},
	author = {He, Zhitao and Cao, Pengfei and Wang, Chenhao and Jin, Zhuoran and Chen, Yubo and Xu, Jiexin and Li, Huaijun and Liu, Kang and Zhao, Jun},
	year = {2024},
	pages = {9399--9416},
}

@article{Hackenburg2024,
	title = {Evaluating the persuasive influence of political microtargeting with large language models},
	volume = {121},
	issn = {0027-8424, 1091-6490},
	doi = {10.1073/pnas.2403116121},
	language = {en},
	number = {24},
	journal = {Proceedings of the National Academy of Sciences},
	author = {Hackenburg, Kobi and Margetts, Helen},
	month = jun,
	year = {2024},
	pages = {e2403116121},
}

@article{Habernal2024,
	title = {Mining legal arguments in court decisions},
	volume = {32},
	issn = {0924-8463, 1572-8382},
	doi = {10.1007/s10506-023-09361-y},
	language = {en},
	number = {3},
	journal = {Artificial Intelligence and Law},
	author = {Habernal, Ivan and Faber, Daniel and Recchia, Nicola and Bretthauer, Sebastian and Gurevych, Iryna and Spiecker Genannt Döhmann, Indra and Burchard, Christoph},
	month = sep,
	year = {2024},
	pages = {1--38},
}

@inproceedings{Grundler2022a,
	address = {Online and in Gyeongju, Republic of Korea},
	title = {Detecting {Arguments} in {CJEU} {Decisions} on {Fiscal} {State} {Aid}},
	url = {https://aclanthology.org/2022.argmining-1.14/},
	booktitle = {Proceedings of the 9th {Workshop} on {Argument} {Mining}},
	publisher = {International Conference on Computational Linguistics},
	author = {Grundler, Giulia and Santin, Piera and Galassi, Andrea and Galli, Federico and Godano, Francesco and Lagioia, Francesca and Palmieri, Elena and Ruggeri, Federico and Sartor, Giovanni and Torroni, Paolo},
	editor = {Lapesa, Gabriella and Schneider, Jodi and Jo, Yohan and Saha, Sougata},
	month = oct,
	year = {2022},
	pages = {143--157},
}

@inproceedings{Giacchetta2025,
	address = {Vienna, Austria},
	title = {Argumentative {Analysis} of {Legal} {Rulings}: {A} {Structured} {Framework} {Using} {Bobbitt}’s {Typology}},
	shorttitle = {Argumentative {Analysis} of {Legal} {Rulings}},
	doi = {10.18653/v1/2025.argmining-1.10},
	language = {en},
	booktitle = {Proceedings of the 12th {Argument} mining {Workshop}},
	publisher = {Association for Computational Linguistics},
	author = {Giacchetta, Carlotta and Bernardi, Raffaella and Montini, Barbara and Staiano, Jacopo and Tomasi, Serena},
	year = {2025},
	pages = {107--115},
}

@article{Dugac2025,
	title = {Classifying legal interpretations using large language models},
	issn = {0924-8463, 1572-8382},
	doi = {10.1007/s10506-025-09447-9},
	language = {en},
	journal = {Artificial Intelligence and Law},
	author = {Dugac, Gaspar and Altwicker, Tilmann},
	month = apr,
	year = {2025},
}

@article{Dahl2024,
	title = {Large {Legal} {Fictions}: {Profiling} {Legal} {Hallucinations} in {Large} {Language} {Models}},
	volume = {16},
	copyright = {https://creativecommons.org/licenses/by-nc/4.0/},
	issn = {2161-7201, 1946-5319},
	shorttitle = {Large {Legal} {Fictions}},
	doi = {10.1093/jla/laae003},
	language = {en},
	number = {1},
	journal = {Journal of Legal Analysis},
	author = {Dahl, Matthew and Magesh, Varun and Suzgun, Mirac and Ho, Daniel E},
	month = jan,
	year = {2024},
	pages = {64--93},
}

@article{Coan2025,
	title = {Artificial {Intelligence} and {Constitutional} {Interpretation}},
	volume = {96},
	url = {https://heinonline.org/HOL/P?h=hein.journals/ucollr96&i=412},
	language = {eng},
	number = {2},
	journal = {University of Colorado Law Review},
	author = {Coan, Andrew and Surden, Harry},
	year = {2025},
	pages = {413--498},
}

@article{Breum2024,
	title = {The {Persuasive} {Power} of {Large} {Language} {Models}},
	volume = {18},
	issn = {2334-0770, 2162-3449},
	doi = {10.1609/icwsm.v18i1.31304},
	journal = {Proceedings of the International AAAI Conference on Web and Social Media},
	author = {Breum, Simon Martin and Egdal, Daniel Vædele and Gram Mortensen, Victor and Møller, Anders Giovanni and Aiello, Luca Maria},
	month = may,
	year = {2024},
	pages = {152--163},
}

@inproceedings{Blair-Stanek2026,
	address = {New York, NY, USA},
	series = {{ICAIL} '25},
	title = {{LLMs} {Provide} {Unstable} {Answers} to {Legal} {Questions}},
	isbn = {979-8-4007-1939-4},
	doi = {10.1145/3769126.3769245},
	booktitle = {Proceedings of the {Twentieth} {International} {Conference} on {Artificial} {Intelligence} and {Law}},
	publisher = {Association for Computing Machinery},
	author = {Blair-Stanek, Andrew and Van Durme, Benjamin},
	month = jan,
	year = {2026},
	pages = {425--429},
}

@inproceedings{Blair-Stanek2023,
	address = {New York, NY, USA},
	series = {{ICAIL} '23},
	title = {Can {GPT}-3 {Perform} {Statutory} {Reasoning}?},
	isbn = {979-8-4007-0197-9},
	doi = {10.1145/3594536.3595163},
	booktitle = {Proceedings of the {Nineteenth} {International} {Conference} on {Artificial} {Intelligence} and {Law}},
	publisher = {Association for Computing Machinery},
	author = {Blair-Stanek, Andrew and Holzenberger, Nils and Van Durme, Benjamin},
	month = sep,
	year = {2023},
	pages = {22--31},
}

@article{Amaya2025,
	title = {Reasoning in {Character}: {Virtue}, {Legal} {Argumentation}, and {Judicial} {Ethics}},
	volume = {28},
	issn = {1386-2820, 1572-8447},
	shorttitle = {Reasoning in {Character}},
	doi = {10.1007/s10677-023-10414-z},
	language = {en},
	number = {3},
	journal = {Ethical Theory and Moral Practice},
	author = {Amaya, Amalia},
	month = jul,
	year = {2025},
	pages = {359--378},
}

@article{Almeida2024,
	title = {Exploring the psychology of {LLMs}’ moral and legal reasoning},
	volume = {333},
	issn = {00043702},
	doi = {10.1016/j.artint.2024.104145},
	language = {en},
	journal = {Artificial Intelligence},
	author = {Almeida, Guilherme F.C.F. and Nunes, José Luiz and Engelmann, Neele and Wiegmann, Alex and Araújo, Marcelo De},
	month = aug,
	year = {2024},
	pages = {104145},
}

@inproceedings{Suttle2026,
	address = {Singapore},
	title = {Persuadability and {LLMs} as {Legal} {Decision} {Tools}},
	url = {https://arxiv.org/abs/2604.26233},
	booktitle = {Proceedings of the 21st {International} {Conference} on {Artificial} {Intelligence} and {Law} ({ICAIL} 2026)},
	author = {Suttle, Oisin and Lillis, David},
	month = jun,
	year = {2026},
	note = {Tex.code: https://github.com//ishnid/ICAIL-2026},
}

@incollection{Zhang2022a,
	address = {Cham},
	series = {Natural {Language} {Processing} and {Information} {Systems}},
	title = {A {Decade} of {Legal} {Argumentation} {Mining}: {Datasets} and {Approaches}},
	isbn = {978-3-031-08473-7},
	doi = {10.1007/978-3-031-08473-7_22},
	publisher = {Springer International Publishing},
	author = {Zhang, Gechuan and Nulty, Paul and Lillis, David},
	translator = {Rosso, Paolo and Basile, Valerio and Martínez, Raquel and Métais, Elisabeth and Meziane, Farid},
	year = {2022},
	pages = {240--252},
}

@misc{Durmus2024,
	title = {Measuring the persuasiveness of language models},
	url = {https://www.anthropic.com/news/measuring-model-persuasiveness},
	author = {Durmus, Esin and Lovitt, Liane and Tamkin, Alex and Ritchie, Stuart and Clark, Jack and Ganguli, Deep},
	month = apr,
	year = {2024},
}

@inproceedings{Surani2025,
	address = {Chicago, IL, USA},
	title = {What is the law? {A} system for statutory research ({STARA}) with large language models},
	booktitle = {20th {International} {Conference} on {Artificial} {Intelligence} and {Law} ({ICAIL} 2025)},
	author = {Surani, Faiz and Gailmard, Lindsey A. and Casasola, Allison and Magesh, Varun and Robitschek, Emily J. and Ho, Daniel E},
	year = {2025},
}

@inproceedings{Tennant2025,
	title = {Moral alignment for {LLM} agents},
	url = {https://openreview.net/forum?id=MeGDmZjUXy},
	booktitle = {The thirteenth international conference on learning representations},
	author = {Tennant, Elizaveta and Hailes, Stephen and Musolesi, Mirco},
	year = {2025},
}

@article{Gardner2001,
	title = {Legal positivism: 5 1/2 myths},
	volume = {46},
	publisher = {HeinOnline},
	author = {Gardner, John},
	year = {2001},
	pages = {199},
}

@article{Kelly1964,
	title = {Audi {Alteram} {Partem};{Note}},
	language = {en},
	journal = {NATURAL LAW FORUM},
	author = {Kelly, John M},
	year = {1964},
}

@book{Cane2011,
	address = {Oxford},
	edition = {5th ed},
	series = {Clarendon {Law} {Series}},
	title = {Administrative {Law}},
	isbn = {978-0-19-101894-7 978-0-19-969232-3},
	language = {eng},
	publisher = {OUP Oxford},
	author = {Cane, Peter},
	year = {2011},
}

\end{document}